\pdfoutput=1

\documentclass[11pt]{article}

\usepackage{acl}

\usepackage{times}
\usepackage{latexsym}
\usepackage{booktabs}
\usepackage{booktabs}
\usepackage{amsfonts}
\usepackage{nicefrac}
\usepackage{microtype}
\usepackage{todonotes}
\usepackage{listings}
\usepackage{algorithm}
\usepackage{algpseudocode}
\usepackage{soul}
\usepackage{tabularx}
\usepackage{pifont}
\usepackage{xcolor}
\usepackage{graphicx}
\usepackage{subcaption}
\usepackage{hyperref}
\usepackage{comment}

\usepackage{colortbl}
\usepackage{xcolor}
\usepackage{makecell}

\usepackage[normalem]{ulem}
\definecolor{green}{RGB}{36, 214, 36}
\definecolor{red}{RGB}{235, 30, 30}
\definecolor{lightredshade}{HTML}{dea9a9}
\definecolor{lightgreenshade}{HTML}{bce3bd}
\definecolor{lightblueshade}{HTML}{cacbe8}
\definecolor{MyDarkBlue}{rgb}{0,0.08,1}
\definecolor{MyDarkGreen}{rgb}{0.02,0.6,0.02}
\definecolor{MyDarkRed}{rgb}{0.8,0.02,0.02}
\definecolor{MyDarkOrange}{rgb}{0.40,0.2,0.02}
\definecolor{MyPurple}{RGB}{111,0,255}
\definecolor{MyRed}{rgb}{1.0,0.0,0.0}
\definecolor{MyGold}{rgb}{0.75,0.6,0.12}
\definecolor{MyDarkgray}{rgb}{0.66, 0.66, 0.66}

\definecolor{MyYellow}{rgb}{254, 246, 170}
\definecolor{MyBlue}{rgb}{170, 217, 251}

\newcommand{\eg}{{\it e.g.}}%
\newcommand{\ie}{{\it i.e.}}%

\newcommand{\cf}{\textsc{Coffee}}

\usepackage[T1]{fontenc}

\usepackage[utf8]{inputenc}

\usepackage{microtype}

\usepackage{hyperref}
\usepackage{multirow}
\usepackage{graphicx}
\usepackage{pdfpages}
\usepackage{url}
\usepackage{xcolor}
\usepackage{epsfig}
\usepackage{adjustbox}
\usepackage{amsfonts}
\usepackage{amsmath}
\usepackage{amssymb}
\usepackage{booktabs} 
\usepackage{comment}
\usepackage{caption}
\usepackage{subcaption}
\usepackage{textcomp}
\usepackage{relsize}
\usepackage{stmaryrd}
\usepackage{bbm}
\usepackage{rotating}
\usepackage{helvet}
\usepackage{courier}
\usepackage{natbib}
\usepackage{cleveref}

\newcommand\cpemoji{\raisebox{-1pt}{\includegraphics[width=1em]{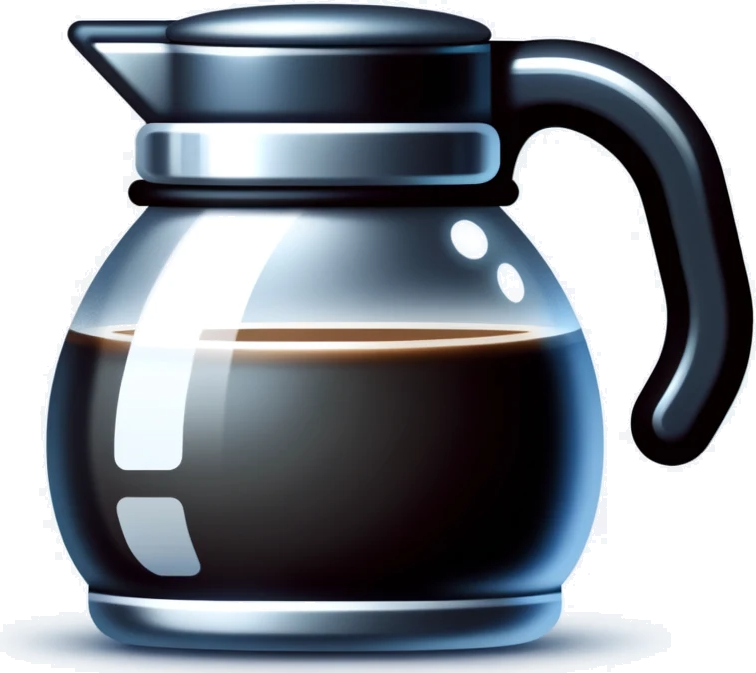}}}
\newcommand\coffee{\raisebox{-2pt}{\includegraphics[width=1em]{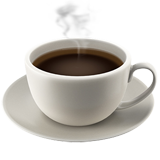}}}

\newcommand\unlocked{\raisebox{-2pt}{\includegraphics[height=1em]{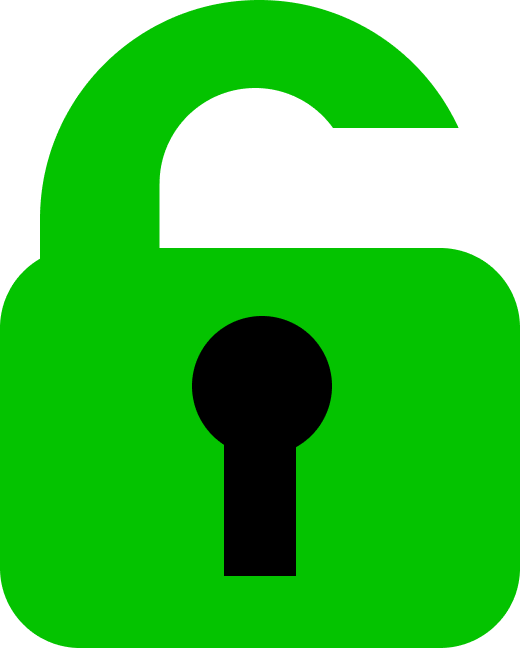}}}
\newcommand\locked{\raisebox{-2pt}{\includegraphics[height=1em]{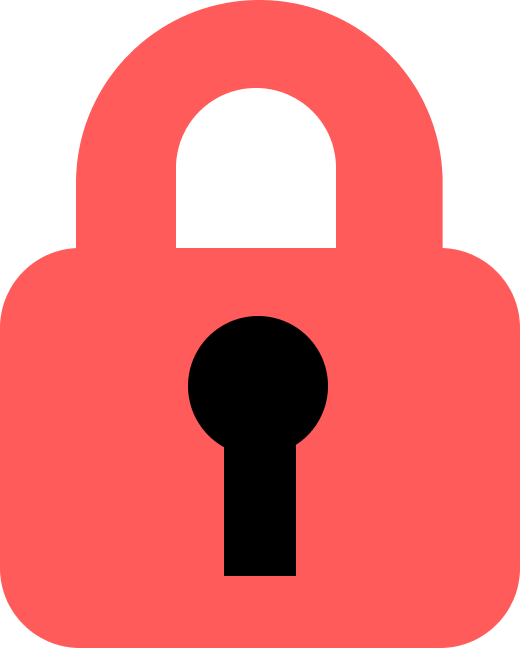}}}
\crefformat{section}{\S#2#1#3} 

\title{\coffee~\textsc{Coffee}: Boost Your Code LLMs by Fixing Bugs with Feedback}
\author{ \qquad Seungjun Moon$^{1}$\thanks{~~Equal contribution} \qquad Hyungjoo Chae$^{1*}$ \qquad \textbf{Yongho Song}$^{1*}$ \qquad \textbf{Taeyoon Kwon}$^{1}$ \\\qquad \textbf{Dongjin Kang}$^{1}$ \qquad \textbf{Kai Tzu-iunn Ong}$^{1}$ \qquad \textbf{Seung-won Hwang}$^{2}$ \qquad \textbf{Jinyoung Yeo}$^{1}$ \\ 
Yonsei University$^{1}$  \quad Seoul National University$^{2}$\\
\texttt{\{lune-blue, mapoout, kopf\_yhs,  jinyeo\}@yonsei.ac.kr} \\ \texttt{seungwonh@snu.ac.kr}\\
}

\begin{document}
\maketitle
\begin{abstract}
    Code editing is an essential step towards reliable programming assistants to automatically correct critical errors in source codes.  
    Recent studies have demonstrated that closed-source LLMs (\ie, ChatGPT and GPT-4) are capable of generating corrective feedback to edit wrong source codes. 
    However, this remains challenging for open-source code LLMs, as they struggle to find critical errors in a source code and suggest correct edits.
    Hence, the focus of our work is to leverage open-source code LLMs to generate helpful feedback with correct guidance for code editing.
    To this end, we present {\textsc{Coffee}}, a dataset specifically designed for code editing with feedback. 
    Along with this dataset, we propose \textsc{CoffeePots}, a novel framework for \textbf{CO}de \textbf{F}ixing with \textbf{FEE}dback via \textbf{P}reference-\textbf{O}ptimized \textbf{T}uning and \textbf{S}election. 
    The proposed framework automatically generates helpful feedback for code editing by aligning feedback generation with correct edits.
    The combination of \textsc{Coffee} and \textsc{CoffeePots} achieves state-of-the-art performance on code editing benchmark. Codes and checkpoints are publicly available.\footnote{https://github.com/Lune-Blue/COFFEE}
\end{abstract}

\section{Introduction}

Thanks to extensive pre-training on code corpora, large language models (LLMs)~\citep{brown2020gpt3} have shown significant success in code-related tasks~\citep{tyers-etal-2023-codex}.
However, these large language models of code, \ie, code LLMs, usually generate source codes that contain bugs and thus lead to undesirable outputs~\citep{ekbal2022adversarial}.
Therefore, the task of \textbf{code editing}, which requires the model to locate error spans in a source code and make correct edits, is an important step towards building reliable programming assistants~\citep{wei2023coeditor,gupta2020repair}.

\begin{figure}[t]
\centering
    \includegraphics[width=0.9\linewidth]{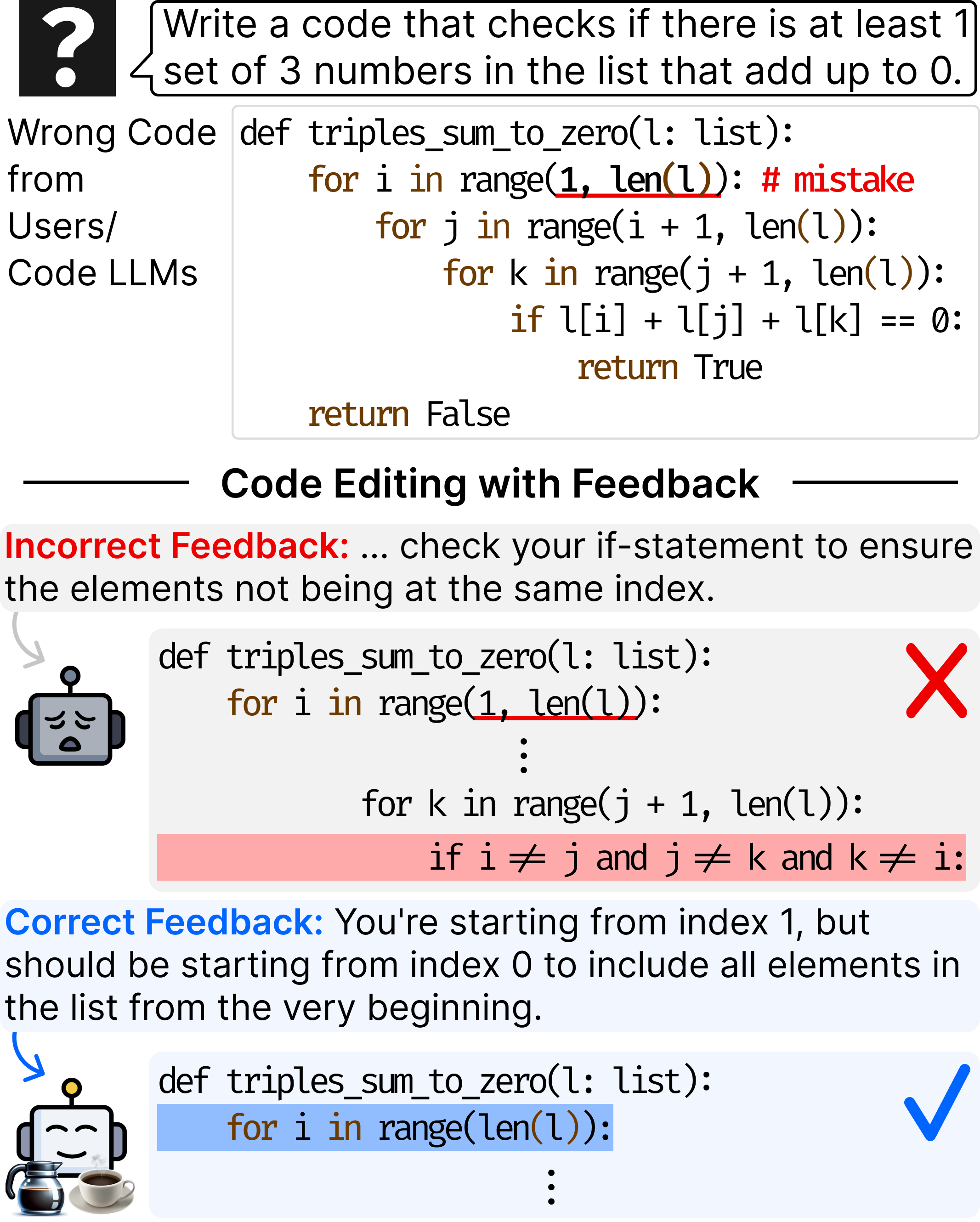}
\caption{A motivating example. The correctness of
feedback affects code LLMs’ performance in editing.} 
\label{fig:motivate}
\end{figure}

Augmenting LLMs with feedback has recently garnered much attention in solving complex code-related tasks~\citep{chen2023teaching,gou2023critic,yang2023intercode}.
In particular, studies have demonstrated the promising capabilities of LLMs to generate feedback for correcting errors~\citep{shinn2023reflexion,zhang2023selfedit}.
However, most of them resort to closed-source LLMs (\eg, ChatGPT and GPT-4~\citep{openai2023chatgpt, openai2023gpt4}) for feedback generation,
which largely limit their real-world applications due to the high API costs and potential security issues~\citep{bommasani2022opportunities}.
Hence, we take a step towards using open-source code LLMs as an alternative to closed-source LLMs for generating helpful feedback that leads to strong code editing performance.

Generating corrective feedback for wrong codes is challenging for open-source code LLMs. 
These models are incapable of capturing small changes in source codes and providing natural language explanations of the codes~\citep{muennighoff2023octopack,micelibarone2023larger}. 
Therefore, the code LLMs tend to generate misleading feedback that does not point to key errors in source codes and suggest incorrect solutions for the given errors, resulting in suboptimal edits as illustrated in Figure~\ref{fig:motivate}. 
The problem arises because open-source code LLMs are trained on code corpora that do not explicitly target the task of code editing. This motivates us to construct a high-quality dataset annotated with natural language feedback for feedback-augmented code editing.

To this end, we present \coffee~\textbf{\textsc{Coffee}}, a dataset curated for \textbf{CO}de \textbf{F}ixing with \textbf{FEE}dback. 
\cf~differs from existing code datasets in the following ways: (1) Our dataset explicitly targets the task of code editing and includes diverse solutions to programming problems that cover various error cases. (2) We additionally annotate natural language feedback that guides the solution of critical errors in an erroneous source code. (3) We leverage ChatGPT to generate synthetic test cases, which are used to assess the correctness of the edited codes and the helpfulness of the corresponding feedback. 

With \cf, we propose \cpemoji ~\textbf{\textsc{CoffeePots}}, a framework for \textbf{CO}de \textbf{F}ixing with \textbf{FEE}dback via \textbf{P}reference-\textbf{O}ptimized \textbf{T}uning and \textbf{S}election. 
Specifically, we leverage \cf~to train open-source code LLMs for feedback-augmented code editing, where a code editing model (\ie, editor) is prompted to generate correct solutions conditioned on the corrective feedback from the feedback generation model (\ie, critic).
To align feedback from the critic with correct edits from the editor, we use the annotated test cases in \cf~to measure the helpfulness of feedback and collect a set of helpful (\ie, preferred) and unhelpful (\ie, dispreferred) feedback. 
This preference set is then used for optimizing the critic model to adhere to helpful solutions, and implementing feedback selector to single out the optimal solution from the critic. 
    

Our contributions are threefold: (1) We present \textbf{\textsc{Coffee}}, a high-quality dataset targeting code editing with feedback; (2) We propose \textbf{\textsc{CoffeePots}}, a framework for code editing with feedback that achieves state-of-the-art (SOTA) performance on code editing benchmark; 
(3) Our extensive analysis on \textbf{\textsc{CoffeePots}} not only demonstrates its efficacy but also lays a strong foundation for future research in the under-explored area of feedback generation with open-source code LLMs.
\section{\coffee~\textsc{Coffee}: A Dataset for Code Fixing with Feedback}
\label{sec:dataset}

\begin{figure}[t]
\centering
    \includegraphics[width=0.92\linewidth]{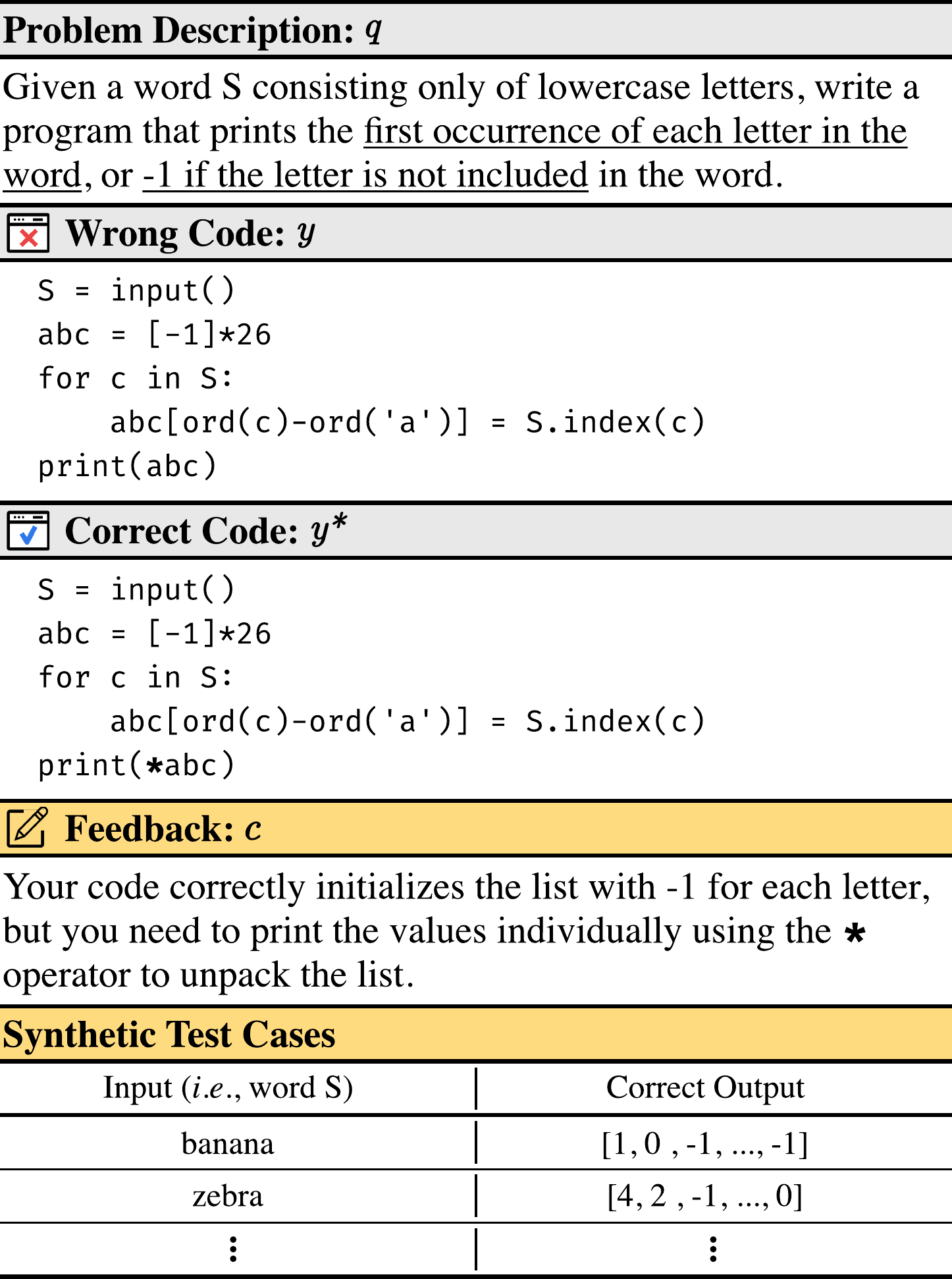}
\caption{An example from \coffee~ \textsc{Coffee}. Elements highlighted in yellow are annotated specifically for feedback-augmented code editing.} 
\label{fig:coffee_example}
\end{figure}

Our motivation is to augment code LLMs with natural language feedback in order to provide useful guidance for code editing. We formulate this as sequence-to-sequence generation, where the model is tasked to generate a correct solution $y^*$ to the problem $q$ conditioned on a wrong source code $y$ and natural language feedback $c$, such that $y^*$ addresses errors in $y$ based on the feedback $c$.

Here, we present \coffee~\textbf{\cf}, a large-scale dataset specifically curated for \textbf{CO}de \textbf{F}ixing with \textbf{FEE}dback. 
Our dataset includes diverse solutions to programming problems collected from an online competitive programming platform. 
For each solution, we additionally annotate natural language feedback to provide detailed explanations for the errors towards correct edits, and augment synthetic test cases to measure the correctness of the edited solutions. We include the statistics of \cf~in Table~\ref{tab:data_statistics} and an example in Figure~\ref{fig:coffee_example}.

\subsection{Collecting Diverse Problems and Solutions}
Training a code editing model requires a large set of correct and wrong solutions (\ie, codes) for diverse programming problems. We achieve this by collecting diverse user submission histories from online competitive programming platforms.
Specifically, for each problem $q$ with a correct submission $y_n$, we collect submission histories $\{y_1, y_2, ..., y_n\}$, where $\{y_k\}_{k=1}^{n-1}$ are incorrect solutions. We then construct $(q,y,y^*)$ triplets by pairing each incorrect solution $y_k$ with the correct one $y_n$, \ie{}, {$\{(q,y_{k}, y_n)\}_{k=1}^{n-1}$}. 

To ensure diversity of difficulty in our dataset, we collect equal numbers of problems for the five difficulty levels provided in the platforms, ranging from beginner to expert levels.
We also make our dataset include various solutions to each problem by collecting submission histories from 100 different users. 
We perform an in-depth analysis to check if there is data leakage from benchmarks we evaluate our model on, and we verify there is no overlap between our data and test data (Appendix~\ref{appendix:overlap}).

\subsection{Annotating Natural Language Feedback}  
We additionally annotate natural language feedback that provides useful guidance on the necessary edits. 
For each triplet $(q,y,y^*)$, we prompt a closed-source LLM (\ie, ChatGPT~\citep{openai2023chatgpt}) to describe how the correct solution $y^*$ differs from the wrong code $y$. 
The resulting description $c$ serves as the feedback that describes necessary changes on the given wrong code $y$ for obtaining the correct code $y^*$. 
To ensure the quality of the annotated feedback, we exclude user submissions that do not involve error correction. 
We discuss details on feedback annotation in Appendix~\ref{appendix:feedback_annotation}, including our ChatGPT prompt and filtering techniques.

\begin{table}[t]
    \centering
    \small
    \begin{tabular}{ll}
    \hline
    \toprule
    \textbf{Statistics} &  \\ 
    \midrule
    \# of instances & 44,782 \\
    \# of total prob. sets & 742 \\
    Avg.~solution len. & 674.1 \\
    Avg.~wrong code len. & 649.4 \\
    Avg.~feedback len. &  269.0 \\
    Avg.~description len. & 573.9 \\
    Avg.~\# of error lines per code & 4.19 \\ 
    Avg.~ \# of submissions per user & 2.7 \\
    Avg.~ \# of hidden test cases per prob. & 35.5 \\
    \bottomrule
    \end{tabular}
    \caption{
        Statistics of \coffee~\textsc{Coffee}.
    }
    \label{tab:data_statistics}
\end{table}



\subsection{Augmenting Synthetic Test Cases}
Finally, we include hidden test cases for each edit instance $(q,y,y^*,c)$ in our dataset to assess whether the edited code is the correct solution to the problem.
As the programming platform does not make test cases publicly available, we annotate synthetic test cases by prompting ChatGPT to generate inputs for a given $q$ and executing them with the correct code $y^*$ to obtain the corresponding outputs.
These synthetic test cases are used to measure the correctness of an edited code and approximate the helpfulness of the feedback, which we later use as supervision signals for training LLMs to generate helpful feedback (\cref{ssec:preference_construct}). More details on synthetic test cases are in Appendix~\ref{appendix:test_case}.




\section{\cpemoji~\textsc{CoffeePots}: Aligning Feedback with Preferred Edits}
\begin{figure*}[ht]
\centering
    \includegraphics[width=1.95\columnwidth]{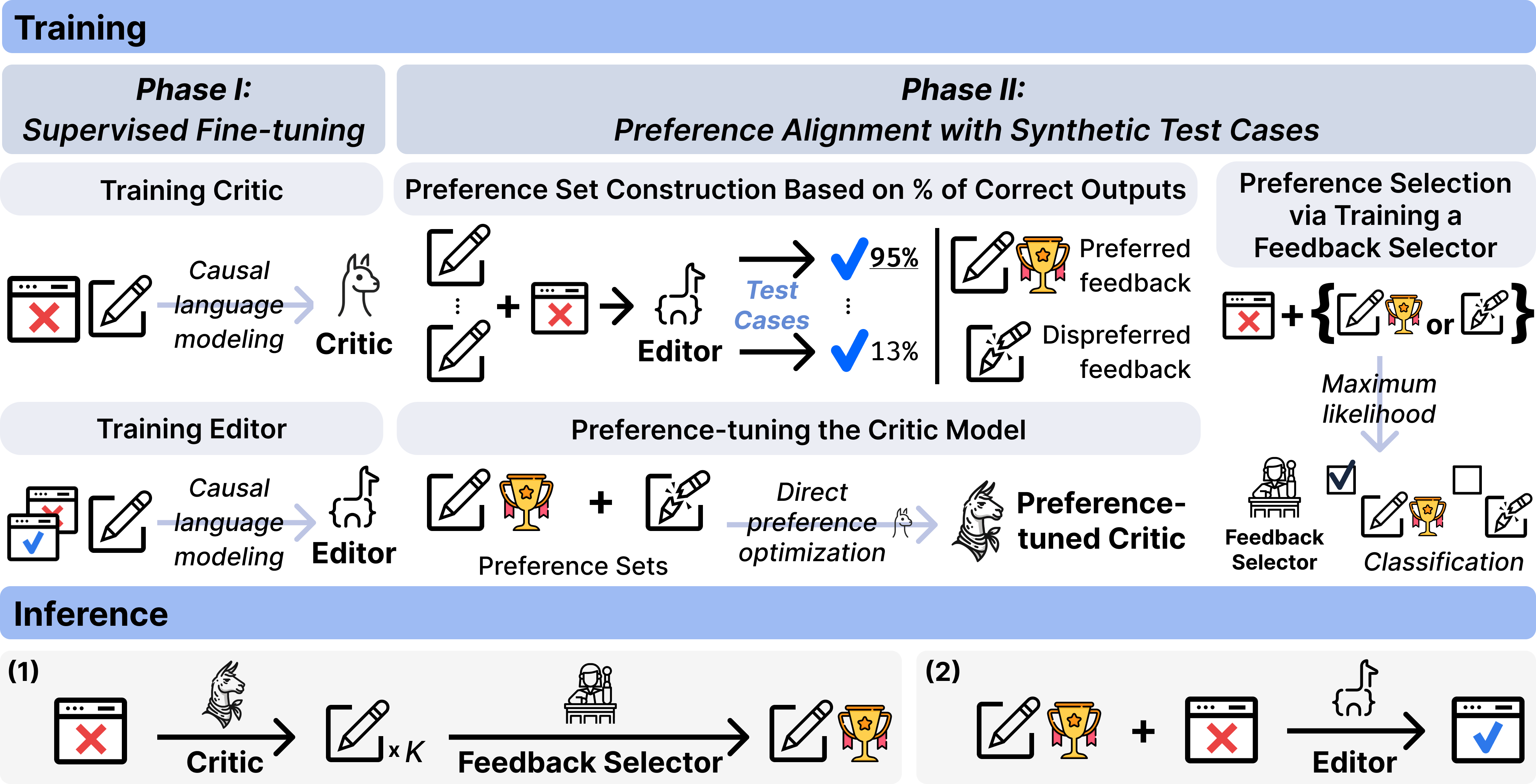}
\caption{The overview of \cpemoji~\textsc{CoffeePots}. }
\label{fig:overview}
\end{figure*}

In this section, we introduce \textsc{CoffeePots}, a framework for \textbf{CO}de \textbf{F}ixing with \textbf{FEE}dback via \textbf{P}reference-\textbf{O}ptimized \textbf{T}uning and \textbf{S}election.
We first use our \cf~dataset to train code LLMs via supervised fine-tuning (SFT) for feedback-augmented code editing.
We then conduct a preliminary study on the effect of feedback on code editing, demonstrating the need for helpful feedback that correctly addresses critical errors. 
Finally, we additionally leverage synthetic test cases in \cf~to annotate preferred (\ie, helpful) solutions and apply preference alignment to guide the generation of helpful feedback.   
Figure~\ref{fig:overview} illustrates the overview of our framework.
\subsection{Phase I: Supervised Fine-tuning for Feedback-augmented Editing}
\label{ssec:sft}

To facilitate code editing with feedback, we fine-tune two open-source code LLMs using \textsc{Coffee} to generate corrective feedback and produce a correct solution, respectively.
Specifically, we employ a \textbf{critic} to generate feedback for code editing and an \textbf{editor} to correct the given wrong code based on the feedback. 
Both models are implemented using the 7B parameter Code {Llama}~\citep{roziere2023code} and are trained via causal language modeling.

\paragraph{Critic.} For each edit instance $(q,y,y^*,c)$ from the \textsc{Coffee} dataset denoted as $\mathcal{D}$, the critic model $\theta$ is trained to predict the correct feedback $c$ for the given problem description $q$ and its incorrect solution $y$:
\begin{equation}\label{eq:critic}
    \mathcal{L}_{\texttt{critic}} = \sum_{(q,y,c) \sim \mathcal{D}} \log p_\theta( c | q, y )
\end{equation}

\paragraph{Editor.} For each $(q,y,y^*,c) \in \mathcal{D}$, the editor model $\phi$ is trained to predict the correct solution $y^*$ conditioned on the correct feedback $c$ as well as the problem description $q$ and the incorrect solution $y$:
\begin{equation}\label{eq:editor}
    \mathcal{L}_{\texttt{editor}} = \sum_{(q,y,y^*,c) \sim \mathcal{D}} \log p_\phi( y^* | q, y, c )
\end{equation}

In practice, we employ QLoRA~\citep{dettmers2023qlora} to efficiently fine-tune our models. For decoding strategies, both models adopt top-$p$ sampling~\citep{holtzman2020nucleus} with $p=0.95$. Implementation details are described in Appendix~\ref{appendix:implementation_details}.  

\begin{figure}[t]
    \centering
    \begin{subfigure}[t]{0.535\linewidth}
        \includegraphics[width=\linewidth]{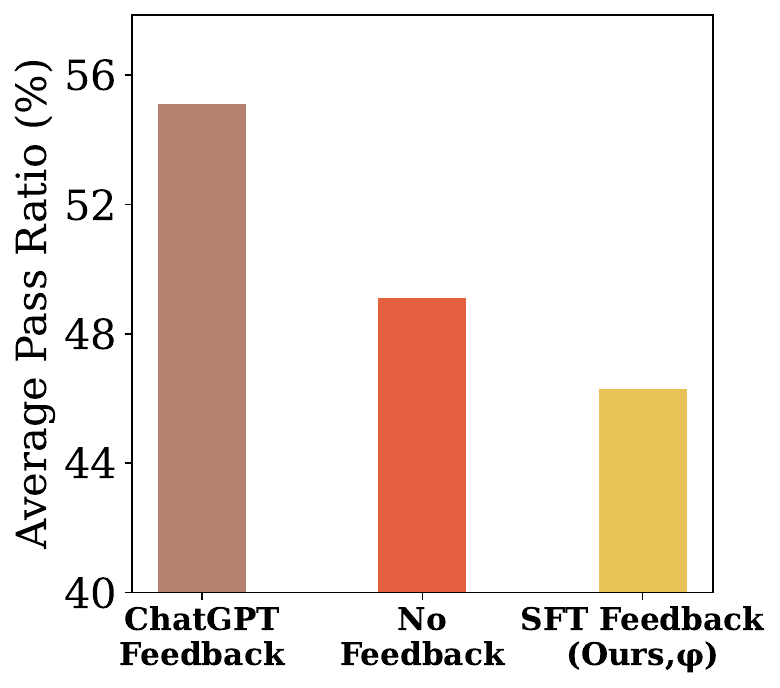}    
        \caption{Performance of critic when given different feedback}
        \label{fig:prelim_pass_ratio}
    \end{subfigure}
    \hfill
    \begin{subfigure}[t]{0.42\linewidth}
        \includegraphics[width=\linewidth]{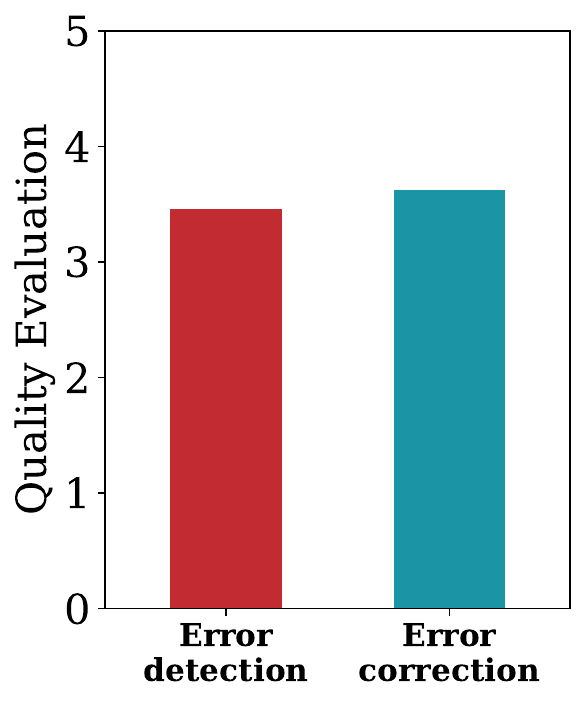}
        \caption{Human evaluation of feedback from the critic $\theta$}   
        \label{fig:prelim_error_analysis}
    \end{subfigure}
    \caption{Preliminary studies on the critic $\theta$.}
\label{fig:pass_ratio}
\end{figure}

\paragraph{Effect of feedback on code editing.}
To study the effect of feedback on code editing, we augment our editor $\phi$ with feedback from our SFT critic $\theta$ and ChatGPT and measure its performance on the test set of \cf. 
Figure~\ref{fig:prelim_pass_ratio} shows a significant performance gain from the editor augmented with feedback from ChatGPT, which demonstrates the benefit of feedback for code editing.\footnote{The performance is measured as the average percentage of test cases where edited solutions produce correct outputs.}


However, we also find that augmenting the editor with feedback from our SFT critic fails to improve the editing performance. 
We conduct human evaluation to assess (1) whether the feedback from our critic correctly identifies errors in a code, and (2) whether it suggests correct edits to the identified errors.
Figure~\ref{fig:prelim_error_analysis} shows that while our SFT critic is capable of providing correct edits to the errors, it struggles to identify critical errors in the given source codes.\footnote{Details on the human evaluation are in Appendix~\ref{appendix:human_error_detection}.}
This motivates us to further align feedback generation with the goal of code editing, such that the generated feedback addresses critical errors in the wrong source codes.
\subsection{Phase II: Preference Alignment with Synthetic Test Cases}

To ensure the alignment between feedback generation and code editing, we leverage synthetic test cases in \cf~to annotate the preference from the editor and guide the framework to produce preferred feedback with correct edits to critical errors. Specifically, we adopt \textbf{preference tuning} and \textbf{selection} as the key strategies for preference alignment. 

\paragraph{Preference set construction.}\label{ssec:preference_construct} We first obtain a preference set of preferred (\ie, helpful) and dispreferred (\ie, unhelpful) feedback using the annotated test cases. 
Given an edit instance $(q,y,y^*,c) \in \mathcal{D}$, we repurpose our critic $\theta$ to sample $K$ feedback candidates $c'_i$ and our editor $\phi$ to generate corresponding candidate solutions $y'_i$.\footnote{We set the number of feedback candidates $K$ as $5$ for our implementation. Ablation on $K$ is shown in Appendix~\ref{app:k_analysis}.}
For each feedback candidate $c'_i$, we leverage the synthetic test cases from \cf~to assign a preference score $s(c'_i)$, which is computed as the proportion of test cases where the solution $y'_i$ produces the correct output. 
The resulting preference set $\mathcal{P}$ consists of preference pairs $(q,y,c^+,c^-)$ containing the \textit{chosen} candidate $c^+$ with the highest preference score and the \textit{reject} candidate $c^-$ with the lowest score.

\paragraph{Preference tuning.}\label{ssec:dpo}
Using the preference set $\mathcal{P}$ constructed using critic $\theta$, we additionally apply Direct Preference Optimization (DPO)~\citep{rafailov2023direct} on the SFT critic model $\theta$ to train a \textit{preference-tuned} critic model $\theta^*$ that reflects the preference of the editor. Formally, the training objective of $\theta^*$ is defined as follows:
\begin{multline}\label{eq:dpo}
    \mathcal{L}_\text{DPO}(\theta^*;\theta) = \\ -\mathop{\mathbb{E}}_{ (q,y,c^+,c^-)\sim \mathcal{P}} \log \sigma \left[ r(q,y,c^+) - r(q,y,c^-) \right]  
\end{multline}
where $\sigma$ denotes the logistic function, and $r$ denotes the reward function on the feedback implicitly defined by $\theta^*$ and $\theta$, with a hyperparameter $\beta$ to control the deviation from $\theta$:\footnote{We follow \citet{rafailov2023direct} and set the hyperparameter $\beta$ as 0.1. See Appendix~\ref{appendix:implementation_details} for details on implementation.}
\begin{equation}
    r(q,y,c) = \beta\log\frac{p_{\theta^*}(c \mid q,y)}{p_{\theta}(c \mid q,y)} 
\end{equation}
By tuning the critic model using preference pairs, the resulting critic model $\theta^*$ is trained to be more biased towards helpful feedback ($c^+$) while avoiding unhelpful feedback ($c^-$).

\paragraph{Preference selection.}\label{ssec:feedback_selector}

Inspired by prior studies on reasoning~\citep{wang2023selfconsistency}, we further design \textsc{CoffeePots} to sample multiple feedback from the preference-tuned critic $\theta^*$ and select the most helpful feedback via a \textit{feedback selector}. 
Specifically, we implement the feedback selector $\psi$ as a binary classifier that classifies the chosen ($c^+$) and rejected ($c^-$) feedback for each instance in the preference set $\mathcal{P}$ constructed using $\theta^*$.
During inference, we sample $K$ diverse feedback candidates from $\theta^*$ using nucleus sampling~\citep{holtzman2020nucleus} and select the feedback $c^*$ with the highest probability from the selector $\psi$.
\begin{equation}
    c^* = \underset{c'_i}{\mathrm{argmax}} \ \log ({p_{\psi}(c'_i | q, y)})
\end{equation}


\section{Experimental Settings}
\subsection{Datasets}
\label{ssec:datasets}

We evaluate the effectiveness of \textsc{CoffeePots} for code editing under two scenarios.
First, we test \textsc{CoffeePots} on a benchmark for \textbf{code editing} with a particular focus on assessing its code editing capability.
For that, we use HumanEvalFix~\citep{muennighoff2023octopack}, which is designed to check whether the models is capable of fixing various types of error in the codes.
Secondly, we further explore whether \textsc{CoffeePots} can boost the \textbf{code generation} performance of LLMs by editing errors in their output codes. 
For that, we apply code editing on the incorrect solution from the generation model for each problem $q$ in the following three code generation benchmarks: 
HumanEvalSynthesize~\citep{muennighoff2023octopack}, APPS~\citep{hendrycksapps2021}, and MBPP~\cite{austin2021program}.


\begin{table}[t]
\small
\renewcommand{\arraystretch}{1.15} 
\centering

\resizebox{\linewidth}{!}{
\begin{tabular}{lcc}
\toprule
\textbf{Methods}&\textbf{Params}&\textbf{Pass@1}\\
\midrule
\cellcolor{gray!17}\locked~~\textit{Closed-source LLMs} & \cellcolor{gray!17}& \cellcolor{gray!17} \\
~~ChatGPT~\citep{openai2023chatgpt} & - & 39.6  \\
~~+Self-Debug& - & 44.5 \\
~~+Self-Refine & - & 45.1 \\
~~GPT-4~\citep{openai2023gpt4} & - & \textbf{47.0}  \\
\midrule
\cellcolor{gray!17}\unlocked~~\textit{Open-source LLMs} &\cellcolor{gray!17} &\cellcolor{gray!17} \\
~~InstructCodeT5+~\citep{wang2023codet5plus} & 16B & 2.7  \\
~~StarChat-$\beta$~\citep{Tunstall2023starchat-alpha} & 16B & 18.1  \\
~~CodeGeeX2~\citep{zheng2023codegeex} & 6B & 15.9  \\
~~StarCoder~\citep{li2023starcoder} & 16B  & 8.7 \\
~~OctoGeeX~\citep{muennighoff2023octopack} & 6B & 28.1 \\
~~OctoCoder~\citep{muennighoff2023octopack} & 16B & 30.4 \\
~~WizardCoder~\citep{luo2023wizardcoder} & 16B & 31.8  \\
~~Code {Llama}~\citep{roziere2023code} & 7B & 15.2  \\
~~Code {Llama}~\citep{roziere2023code} & 13B & 16.4  \\
~~Self-Debug (Code {Llama}) & 13B & 18.9  \\
~~Self-Refine (Code {Llama}) & 13B & 20.1  \\
~~\textsc{CoffeePots} (Ours) & \textbf{7B} & \textbf{51.2}  \\


\bottomrule
\end{tabular}
}

\caption{Code editing results on HumanEvalFix. Baseline performances are from \citet{muennighoff2023octopack}.}
\label{tab:humanevalfix}
\end{table}





\begin{table*}[t]
\small
\renewcommand{\arraystretch}{1.15} 
\centering

\resizebox{\textwidth}{!}{
\begin{tabular}{lccccc}
\toprule
\multirow{2}{*}{\textbf{Methods}} & \multirow{2}{*}{\textbf{HumanEvalSynthesize}} & \multirow{2}{*}{\textbf{MBPP}}  & \multicolumn{3}{c}{\textbf{APPS}} \\
\cline{4-6}
& & & Introductory & Interview & Competition \\

\midrule
Code generation w/o editing (ChatGPT) &67.6 &49.0 & 48.4 & 21.0 & 5.7 \\
\midrule
\midrule
\cellcolor{gray!17}\locked~~\textit{Editing with closed-source LLM} &\cellcolor{gray!17}&\cellcolor{gray!17}&\cellcolor{gray!17}&\cellcolor{gray!17}& \cellcolor{gray!17}\\
~~~~~ChatGPT &\textbf{74.3 (20.7\%)} &\textbf{52.4 (6.7\%)}  &\textbf{51.9 (6.8\%)}  &\textbf{22.6 (2.0\%)}  &6.0
(0.3\%) \\
~~~~~+ Self-Debug (ChatGPT) &72.5 (15.1\%) &51.2 (4.3\%)  &50.6 (4.3\%)  &22.2 (1.5\%)  &\textbf{6.9 (1.3\%)} \\
~~~~~+ Self-Refine (ChatGPT) &72.5 (15.1\%) &50.4 (2.7\%)  &49.1 (1.4\%)  &21.4 (0.5\%)  &5.9 (0.2\%) \\
\midrule
\cellcolor{gray!17}\unlocked~~\textit{Editing with open-source LLM} &\cellcolor{gray!17}&\cellcolor{gray!17}&\cellcolor{gray!17}&\cellcolor{gray!17}& \cellcolor{gray!17}\\
~~~~~OctoCoder (16B) &68.2 (1.9\%) &49.6 (1.2\%)  &48.4 (0.0\%)  &21.1 (0.1\%)  &5.7 (0.0\%) \\
~~~~~StarCoder (16B) &67.6 (0.0\%) &49.0 (0.0\%)  &48.4 (0.0\%)  &21.0 (0.0\%)  &5.7 (0.0\%) \\
~~~~~WizardCoder (16B) &69.5 (5.9\%) &49.8 (1.6\%)  &48.4 (0.0\%)  &21.1 (0.1\%)  &5.7 (0.0\%) \\
~~~~~Code Llama (7B) &70.1 (7.7\%) &49.4 (0.8\%)  &48.4 (0.0\%)  &21.1 (0.1\%)  &5.7 (0.0\%) \\
~~~~~Code Llama (13B) &71.9 (13.3\%) &49.4 (0.8\%)  &48.5 (0.2\%)  &21.1 (0.1\%)  &5.7 (0.0\%) \\
~~~~~Self-Debug (Code Llama 13B) &70.7 (9.6\%) &50.0 (2.0\%)  &48.5 (0.2\%)  &21.0 (0.0\%)  &5.7 (0.0\%) \\
~~~~~Self-Refine (Code Llama 13B) &73.7 (18.8\%) &50.6 (3.1\%)  &48.6 (0.4\%)  &21.5 (0.6\%)  &5.8 (0.1\%) \\
~~~~~\textsc{CoffeePots} (7B) &\textbf{75.0 (22.8\%)} &\textbf{52.8 (7.5\%)}  &\textbf{49.3 (1.7\%)}  &\textbf{21.8 (1.0\%)}  &\textbf{6.4 (0.7\%)} \\
\bottomrule
\end{tabular}
}

\caption{Performances in editing machine-generated codes. We report pass@1 and ERR (in parentheses). We use ChatGPT (the first row) to generate codes for problems from several benchmark datasets for code generation.}
\label{tab:codegen}
\end{table*}
\subsection{Metrics}
Following \citet{muennighoff2023octopack}, we report the code editing performance using \textbf{pass@1}, which is the percentage of problems in the test set that the model solves in a single attempt.
For code generation benchmarks, we additionally measure \textbf{Error-Resolved Rate (ERR)} as the proportion of correct edits among a subset of problems with incorrect solutions from the generation model. 
This new metric is introduced to compensate for the difference in evaluation methods between two experiment settings: Unlike code editing benchmarks where all solutions are considered for editing, we only edit solutions that are found to be erroneous in code generation benchmarks (\eg{}, 51\% in MBPP).\footnote{Note that ERR is equivalent to pass@1 in code editing benchmarks since all problems are annotated with incorrect solutions. See Appendix~\ref{appendix:benchmarks} for more details.}

\subsection{Baselines}
We choose both \textbf{open-source} and \textbf{closed-source} LLMs as our baselines. For open-source baselines, we use popular foundation models for code-related tasks, including StarCoder~\citep{li2023starcoder}, WizardCoder~\citep{luo2023wizardcoder}, OctoCoder~\citep{muennighoff2023octopack}, and CodeLlama family~\citep{roziere2023code}.
We also consider two closed-source baselines, ChatGPT~\citep{openai2023chatgpt} and GPT-4~\citep{openai2023gpt4}. Our hypothesis is that \textsc{CoffeePots} enhances the code editing capabilities of open-source models and shows performance competitive to the closed-source baselines.

We further consider two prominent prompting methods to implement baselines that incorporate feedback into code-related tasks: (1) Self-Debug~\citep{chen2023teaching} is a method specialized in fixing bugs by incorporating \textbf{compiler feedback} (\eg{}, unit test results).
(2) Self-Refine~\citep{madaan2023selfrefine} uses \textbf{natural language feedback} to iteratively refine models' outputs.
\section{Experimental Results}
\subsection{Performance in Code Editing}

\paragraph{Code editing benchmark.}
Table~\ref{tab:humanevalfix} compares the code editing performance of different models on HumanEvalFix.
We observe that \textsc{CoffeePots} shows significantly higher pass@1 than all open-source baselines, including those with larger parameter sizes.
Moreover, \textsc{CoffeePots} even outperforms the closed-source LLMs (\ie, ChatGPT and GPT-4), achieving SOTA performance on the HumanEvalFix benchmark. 
Overall, \textsc{CoffeePots} exhibits strong capabilities in editing erroneous codes by generating helpful feedback. 

\paragraph{Code generation benchmark.}
Table~\ref{tab:codegen} reports the model performance in editing solutions generated from ChatGPT for problems in HumanEvalSynthesize, MBPP, and APPS. 
\textsc{CoffeePots} outperforms all open-source baselines, including Code Llama (13B), the previous SOTA among open-source code LLMs.
Furthermore, \textsc{CoffeePots} shows better results than feedback-augmented Code Llama (13B), \ie{}, prompted with Self-Refine and Self-Debug, suggesting the effectiveness of our strategy on generating feedback.
In addition, while some open-source code LLMs show almost no improvement in MBPP and APPS (\ie{}, 0\% ERR), \textsc{CoffeePots} shows moderate improvements on these benchmarks (\ie{}, up to 7.5\% ERR).
Compared to closed-source baselines (\ie{}, ChatGPT), \textsc{CoffeePots} achieves competitive results particularly on HumanEvalSynthesize and MBPP, showing that our framework can serve as a strong alternative to closed-source LLMs while being publicly available and much smaller in size. 
\begin{table}[t]
    \centering
    
    \resizebox{\linewidth}{!}{
    \begin{tabular}{lc|ccc}
    \toprule
         & \textbf{H.E.Fix} & \textbf{H.E.Synth} & \textbf{MBPP} & \textbf{APPS} \\
         \midrule
       Error detection  & 3.79& 3.58& 3.50& 3.48\\
       ERR &39.6\%& 20.7\% & 6.7\% & 2.0\% \\
    \bottomrule
    \end{tabular}
    }
    \caption{Human evaluation on the correctness of feedback  on two different experimental settings (denoted as Error detection), combined with ERR of ChatGPT.}
    \label{tab:error_detection_and_err}
\end{table}


\begin{table}[t]
\small
\renewcommand{\arraystretch}{1.15} 
\centering

\begin{tabular}{lc}
\toprule
\textbf{Methods}&\textbf{\% Errors found}\\
\midrule
\cellcolor{gray!13}\locked~~\textit{Closed-source LLMs} &  \cellcolor{gray!13}\\
~~~~~ChatGPT  & 42.1    \\
~~~~~GPT-4 & 48.2  \\
\midrule
\cellcolor{gray!13}\unlocked~~\textit{Open-source LLMs} & \cellcolor{gray!13}  \\
~~~~~OctoGeeX & 26.7 \\
~~~~~OctoCoder  & 36.9 \\
~~~~~WizardCoder & 27.7 \\
~~~~~\textsc{CoffeePots} (Ours) & \textbf{53.3}  \\

\bottomrule
\end{tabular}

\caption{Average proportions of error lines found by different code editing models on HumanEvalFix.}
\label{tab:line_accuracy}
\end{table}

\begin{table}[t]
\small
\renewcommand{\arraystretch}{1.15} 
\centering
\resizebox{\linewidth}{!}{

    \begin{tabular}{lccc}
        \toprule
        
        \multirow{2}{*}{\textbf{Error types}} & {GPT-4} & {OctoCoder} & \textsc{CoffeePots} \\
         & \locked & \unlocked & (Ours)~\unlocked \\
        \midrule
        Missing logic & \textbf{45.5} & 31.2 & 42.4 \\
        Excess logic & 38.7 & 11.0 & \textbf{51.6} \\
        Value misuse & 50.0 & 45.1 & \textbf{54.5} \\
        Operator misuse & 56.0 & 34.4 & \textbf{68.0}  \\
        Variable misuse &\textbf{43.5} & 30.4 & \textbf{43.5}  \\
        Function misuse &\textbf{50.0} & 37.5 & 37.5  \\
        \midrule
        \textbf{Total}& 47.0 & 31.8 & \textbf{51.2} \\
        \bottomrule
    \end{tabular}
}    
    \caption{
        Breakdown of pass@1 on HumanEvalFix by different error types. The performances of the baselines are reported in \citet{muennighoff2023octopack}. 
    }
    \label{table:error_type}
\end{table}

\paragraph{Discrepancy between code editing and generation.}
To better understand the different results from code editing and generation benchmarks, we conduct human evaluation that assesses the correctness of feedback from ChatGPT on erroneous codes.
Specifically, we task the raters to rate how well ChatGPT feedback addresses errors in the solutions on a Likert scale of 1-5.\footnote{Details on the human evaluation are in Appendix~\ref{appendix:human_error_detection}.}
In Table~\ref{tab:error_detection_and_err}, we observe that compared to code editing benchmarks, feedback from ChatGPT is less accurate in finding errors from solutions on the code generation benchmarks. This demonstrates that it is more challenging to identify errors in solutions from code LLMs on code generation benchmarks, resulting in lower ERR for both ChatGPT and \textsc{CoffeePots}. 

\subsection{Analyses on the Effect of Feedback}
\paragraph{Error detection.}
To examine whether feedback from \textsc{CoffeePots} helps the editor better detect errors, we measure how well the code editing models locate critical errors in a source code with and without feedback. 
Specifically, we compute the percentage of errors as the number of edited error lines divided by the number of total error lines in the wrong code. 
In Table~\ref{tab:line_accuracy}, we observe that \textsc{CoffeePots} shows a significantly higher rate of errors found than the baselines on HumanEvalFix, suggesting that augmenting the editor with feedback benefits the model in the localization of errors.

\paragraph{Error types.} Table~\ref{table:error_type} breaks down the performance of open- and closed-source LLMs on HumanEvalFix by different error types. 
We observe that \textsc{CoffeePots} shows particularly strong performance on problems with excess logic or value/operator misuse, which require models to locate the exact error spans. 
We also find that \textsc{CoffeePots} achieves performance comparable to GPT-4 on problems with missing logic, indicating the helpfulness of feedback from our critic in understanding the underlying logic in source codes.
Overall, \textsc{CoffeePots} consistently surpasses open-source baselines for all error types. We include full results with more baselines in Appendix~\ref{appendix:error_types}.

\subsection{Ablation Studies}\label{analysis:ablation_study}
We conduct ablation studies to investigate the effectiveness of each component in \textsc{CoffeePots}. Here, we consider the following ablations: (1) \textbf{Supervised Fine-tuning (SFT)}: a vanilla critic model trained with only SFT (\cref{ssec:sft}). (2) \textbf{Preference tuning}: a vanilla (SFT) critic that further goes through preference-tuning (\cref{ssec:dpo}) without feedback selection. (3) \textbf{Feedback selection}: a vanilla (SFT) critic model paired with a feedback selector (\cref{ssec:feedback_selector}) that selects the best feedback candidates from the critic. 

\begin{table}[t]
\small
\renewcommand{\arraystretch}{1.15} 
\centering
        {
            \begin{tabular}{lcc}
                \toprule
                \textbf{Methods} & \textbf{Pass@1} \\
                \midrule
                Editor w/o feedback ~~~~~~~~~~ & 42.6 \\ 
                ~~+ Preference tuning on editor (\cref{ssec:dpo}) ~~~~~~~~~ & 45.7 \\
                \midrule
                Supervised fine-tuning (\cref{ssec:sft}) ~~~~~~~~~~ & 38.4 \\
                ~~+ Preference tuning on critic (\cref{ssec:dpo}) ~~~~~~~~~ & 45.9 \\
                ~~+ Feedback selection (\cref{ssec:feedback_selector}) ~~~~~~~~~~ & 45.1 \\
                \textsc{CoffeePots} (Ours) ~~~~~~~~~~ & \textbf{51.2} \\
                
                \bottomrule
            \end{tabular}
        }
        \caption{Results of ablation studies on HumanEvalFix.}
        \label{table:ablation}
    \end{table}
\paragraph{Preference alignment benefits code editing.}\label{analysis:helpful_feedback}
Table~\ref{table:ablation} compares the pass@1 performance on the HumanEvalFix benchmark of each setting.
We see a decrease in code editing performance when the editor is provided with feedback from the SFT critic model, 
which is in line with the finding that critic trained with only SFT fails to generate feedback for correct solutions. 
On the contrary, we observe that applying preference tuning and feedback selection on the SFT critic model leads to significant performance gain from the editor, and that \textsc{CoffeePots} shows a significant performance gap between the preference-tuned editor, demonstrating the effectiveness of our approach. 

\begin{figure}[t]
\centering
    \includegraphics[width=\linewidth]{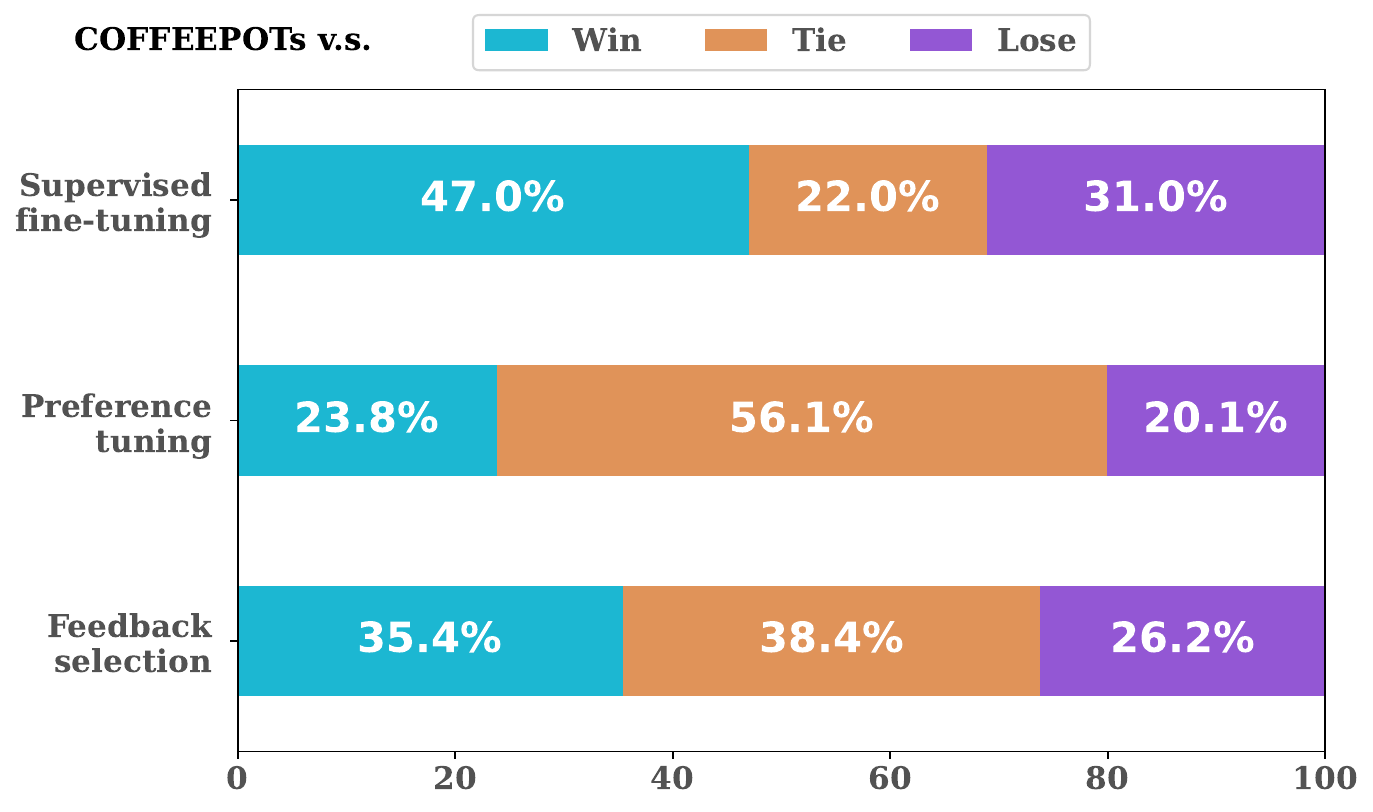}
\caption{Evaluation on the quality of the feedback from \textsc{CoffeePots} and its variants.}
\label{fig:g-eval}
\end{figure}

\paragraph{Preference alignment improves the quality of feedback.}
We conjecture that aligning feedback with the preference of the editor largely improves its quality. To validate this, we follow \citet{liu2023geval} and prompt GPT-4 to conduct pair-wise comparison on feedback from \textsc{CoffeePots}
and its variants using the prompt in Table~\ref{tab:gpt4_compare_prompt}. 
Figure~\ref{fig:g-eval} shows that for 47$\%$ of the test cases, \textsc{CoffeePots} produces feedback with higher quality than the SFT critic. 
Notably, we find that applying preference tuning largely improves the likelihood of generating high-quality feedback, as the win percentage of \textsc{CoffeePots} decreases from 47.0$\%$ to 23.8$\%$. 

\begin{figure}[t]
\centering
    \includegraphics[width=\linewidth]{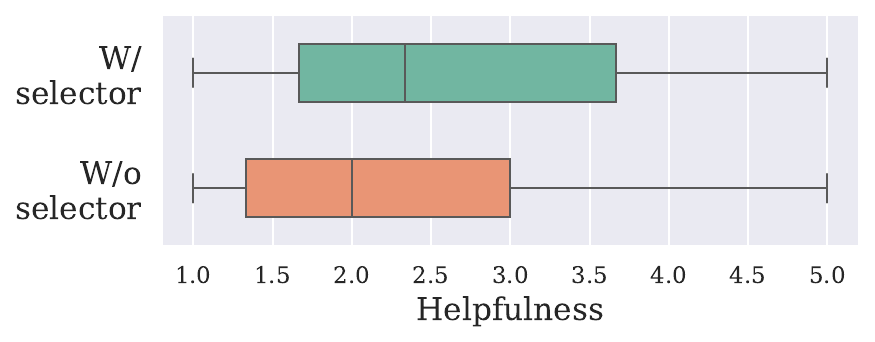}
\caption{Evaluation results between the quality of feedback from \textsc{CoffeePots} with and without selector. We measure the quality of the feedback using GPT-4. }
\label{fig:feedback_quality}
\end{figure}
\paragraph{Feedback selector leaves out unhelpful feedback.}
We observe in Figure~\ref{fig:g-eval} that applying feedback selection on the SFT critic improves the feedback quality. To see how feedback quality benefits from feedback selection, we measure the helpfulness of feedback from \textsc{CoffeePots} and the preference-tuned critic via GPT-4 and compare their distribution in Figure~\ref{fig:feedback_quality}.\footnote{See Appendix~\ref{appendix:feedback_quality_eval} for details on quality evaluation.} 
We see that applying our feedback selector to the preference-tuned critic (\ie, \textsc{CoffeePots}) effectively reduces low-quality feedback from the critic, largely improving the overall quality of the generated feedback.
\section{Related Work}
\paragraph{Code-related tasks with LLMs.}
Pre-training with massive code corpora has been a popular practice for enabling LLMs to generate codes for natural language problems. For instance, \citet{chen2021codex} introduce Codex, which is trained on large-scale Python source codes; \citet{roziere2023code} and \citet{wang2023codet5plus} further introduce instruction-tuning to code LLMs for better code generation.
Yet, it remains challenging to leverage code LLMs for editing critical errors in programs~\citep{li2022competition}.

\paragraph{Code editing with LLMs.}
The recent success of LLMs has spurred an interest in leveraging LLMs for automatic program repair~\citep{fan2023autorepair}. 
An intuitive approach to improve code editing is to augment code editing models with feedback from off-the-shelf programming tools (\eg, compilers)~\citep{gou2023critic,chen2023teaching}. 
Inspired by recent efforts on generating reflective feedback~\citep{madaan2023selfrefine, shinn2023reflexion},  
\citet{zhang2023selfedit} further employ closed-source LLMs to generate natural language explanations on errors.
Our work builds upon these studies and proposes an effective framework for feedback-augmented code editing with open-source LLMs, mitigating high API cost and security issues from closed-source LLMs.

\section{Conclusion}
This paper aims to facilitate feedback-augmented code editing with open-source code LLMs. 
To this end, we present \textsc{Coffee}, a dataset for code fixing annotated with natural language feedback. 
We also propose \textsc{CoffeePots}, a framework for feedback-augmented code editing that aligns feedback generation with correct edits from the editor.
Through extensive experiments, we demonstrate that \textsc{CoffeePots} largely enhances the code editing capabilities of open-source LLMs and achieves SOTA performance on the code editing benchmark by providing the editor with helpful feedback.

\section*{Limitations}\label{sec:limitations}

\paragraph{Scope of editing.}
\textsc{CoffeePots} is trained to edit only erroneous codes according to our goal of correcting errors. Future work might delve into improving readability of codes via editing.
Also, we mainly focus on editing incorrect source codes in a competitive programming setting, which might be slightly far from real-world scenarios (\eg, drawing a bar plot using matplotlib). We assume that further applying other instruction tuning methods~\citep{muennighoff2023octopack, köpf2023oasst} or training on general code corpus~\citep{li2023starcoder} could expand the capability of \textsc{CoffeePots} to general domains.

\paragraph{Using synthetic test cases for measuring reward.}
While running synthetic test cases and using the resulting pass rates might be a promising proxy for calculating reward in preference tuning, there might be edge cases where even erroneous codes pass the synthetic test cases. Further research can incorporate \citet{liu2023is} to make more challenging test cases that can rigorously identify erroneous codes without missing edge cases.

\paragraph{Single programming language.}
Our implementation of \textsc{CoffeePots} is limited to a single programming language, \ie{}, Python. However, future work might apply a similar strategy as ours to expand our model to a multilingual setting, where the model is capable of understanding and editing diverse programming languages.

\paragraph{Single parameter size and architecture.}
Lastly, we implement the critic and editor models only with one parameter size and architecture. However, future work can apply our method to models with larger parameter sizes (\eg{}, Code Llama 70B), which is expected to perform better in code editing. Our framework can also be further applied to other architectures, as our method is model-agnostic.


\section*{Ethical Considerations}
While our dataset originates from online competitive programming platforms, we have ensured the exclusion of personal information to maintain privacy standards.
Additionally, we are aware of the potential risks associated with texts generated by language models, which can contain harmful, biased, or offensive content.
However, based on our assessments, this risk is mostly mitigated in our work.
Lastly, there exists a risk of hallucination in the process of feedback generation and code editing, leading to incorrect edits. 
This emphasizes the need for careful application in our approach.


\bibliography{anthology,custom}

\appendix

\appendix
\clearpage
\renewcommand{\thesection}{\Alph{section}}


\section{Details of \coffee~\textsc{Coffee}}\label{appendix:dataset}

\subsection{Feedback Annotation}\label{appendix:feedback_annotation}
We use \texttt{gpt-3.5-turbo-0613} among the available ChatGPT models to annotate feedback for our dataset. We apply top-$p$ sampling and temperature, where $p = 0.95$ and $T = 0.7$. We limit the number of generation tokens to 500. We leave out submission histories where the LLM fails to find any errors. We also filter out submissions from different users whose correct solutions are identical, as these solutions are usually copied from the web without undergoing editing processes. The prompt used for feedback annotation is in Table~\ref{tab:prompt-annotation}.

\subsection{Synthesizing Test Cases}\label{appendix:test_case}
We prompt ChatGPT to synthesize test cases given a problem description with three demonstrations. We provide the prompt in Table~\ref{tab:synthetic_test_case}.

\section{Implementation Details}\label{appendix:implementation_details}
All components of our framework are implemented using Code {Llama} 7B~\citep{roziere2023code} as the backbone model. For training, we employ QLoRA~\citep{dettmers2023qlora}, incorporating 4-bit quantization with a learning rate of 2e-4 and a batch size of 4. The training is run on 8 NVIDIA GeForce RTX 3090 GPUs. 
Regarding the LoRA configuration, we specify the dimension of low-rank metrices as 64, and alpha as 16. We train over a maximum of 5 epochs, and select the checkpoint based on the validation loss.

\subsection{Phase I: Supervised Fine-tuning for Feedback-augmented Editing}
\paragraph{Critic.}
For critic, we fine-tune Code {Llama} 7B to predict the feedback $c$ given a problem description $q$ of a problem and an incorrect solution $y$. The format of inputs/outputs for training and inference is presented in Table~\ref{tab:critic_prompt}.

\paragraph{Editor.}
For editor, we fine-tune Code {Llama} 7B to predict the correct code $y^*$ conditioned on the correct feedback $c$ as well as the problem description $q$ and the incorrect solution $y$.
The format of inputs/outputs for training and inference is presented in Table~\ref{tab:editor_prompt}.

\subsection{Phase II: Preference Alignment with Synthetic Test Cases}
We first obtain a preference set of preferred and dispreferred feedback using the annotated test cases.

\paragraph{Preference Tuning.}
Given a problem description, a wrong code, and the corresponding preference set, we apply Direct Preference Optimization (DPO)~\citep{rafailov2023direct} to train our critic. That is, we tune critic model to be biased towards helpful feedback. We use the prompt in Table~\ref{tab:critic_prompt} to apply DPO to critic.

\paragraph{Feedback Selector.}
To train the feedback selector with preference sets, we add a classification layer on top of Code {Llama}.
Specifically, given the problem description and the wrong code (two are concatenated) and one feedback from the preference set, we train the feedback selector to classify whether the given feedback is a preferred feed.

\paragraph{Terms and License.} For our implementation and evaluation, we use Huggingface library and vLLM library.\footnote{\url{https://huggingface.co/}} Both libraries are licensed under Apache License, Version 2.0.
We have confirmed that all of the artifacts used in this paper are available for non-commercial scientific use.

\section{Experimental Details}\label{appendix:experimental_details}
\subsection{Baselines}\label{appendix:baselines}
For our experiments, we consider the following open-source baselines:

\paragraph{Code {Llama}.}
Code Llama~\citep{roziere2023code} refers to variants of LLaMA2, specialized in code domains via fine-tuning on code corpus. 
This collection includes various models tailored for specific uses: the foundation model (Code Llama), a Python-focused model (Code Llama-Python), and an instruction-following model (Code Llama-Instruct). 
These models are available in sizes of 7B, 13B, and 34B parameters. 
In our experiments, we use the Code Llama-Instruct model as the baseline.

\paragraph{StarCoder.}
StarCoder~\citep{li2023starcoder} is a code LLM with 15.5B parameters, developed by the BigCode community. 
Based on StarCoderBase, StarCoder is fine-tuned on a corpus with 35 billion Python tokens.
The model checkpoint of StarCoder is publicly available, emphasizing safety and responsible use.

\paragraph{WizardCoder.}
\citet{luo2023wizardcoder} enhance code LLMs by incorporating complex instruction fine-tuning using the Evol-Instruct method to the code domain. 
WizardCoder shows superior performance in popular code generation benchmarks such as HumanEval, HumanEval+, MBPP, and DS1000, outstripping other open-source code LLMs and even surpassing leading closed models in certain evaluations.

\paragraph{InstructCodeT5+.}
InstructCodeT5+~\citep{wang2023codet5plus} is an advanced encoder-decoder LLM tailored for code tasks, overcoming limitations of previous models in architecture and pre-training. 
It offers flexible module combinations for diverse code applications, enhanced by a mix of pre-training objectives like span denoising and text-code matching.

\paragraph{CodeGeeX2.}
CodeGeeX2~\citep{zheng2023codegeex} is a multilingual code generation model with 13B parameter that excels in code generation and translation. 
It is pre-trained on 850 billion tokens from 23 languages and surpasses similar models in HumanEval-X benchmarks. 
CodeGeeX2 is integrated into major coding platforms, enhancing user coding efficiency significantly.

\paragraph{OctoCoder.}
\citet{muennighoff2023octopack} propose OctoCoder, which is trained to follow instructions on codes.
They collect CommitPack, which encompasses an extensive collection of 4 terabytes of Git commits from over 350 programming languages.
They also use OASST~\citep{köpf2023oasst} for instruction tuning. 
\citet{muennighoff2023octopack} demonstrate that instruction tuning plays a critical role in the performance of OctoCoder in diverse code-related tasks.

\paragraph{OctoGeeX.}
Using the same datasets as OctoCoder, \citet{muennighoff2023octopack} train OctoGeeX based on CodeGeeX2 6B.

\paragraph{StarChat-$\beta$.}
StarChat-$\beta$~\citep{Tunstall2023starchat-alpha} is a variant of StarCoder fine-tuned from StarCoderPlus using an uncensored openassistant-guanaco dataset. 
This approach enhances its coding task performance, although it may produce problematic text, positioning it mainly for educational and research use.

For closed-source baselines we consider ChatGPT~\citep{openai2023chatgpt} and GPT-4~\citep{openai2023gpt4}. Note that GPT-4 has shown the SOTA performance in code-related tasks. As these models are sensitive to input prompt, we use the prompts used in \citet{muennighoff2023octopack} to evaluate thes models.

\begin{table*}[!ht]
\small
\renewcommand{\arraystretch}{1.05} 
\centering
\resizebox{0.98\textwidth}{!}{

    \begin{tabular}{lccccccc}
        \toprule
        
        \multirow{2}{*}{\textbf{Error types}}& \multicolumn{2}{c}{\locked~ \textbf{Closed-source}} & &\multicolumn{4}{c}{\unlocked~ \textbf{Open-source}} \\ 
        \cmidrule{2-3} \cmidrule{5-8}
         & ChatGPT & GPT-4 &  & OctoGeeX & OctoCoder & WizardCoder  & \textsc{CoffeePots} (Ours)  \\
        \midrule
        Missing logic (20.1 \%)&36.4 & \textbf{45.5} & & 24.2& 24.4 & 31.2  & 42.4 \\
        Excess logic (18.9 \%)&29.0 & 38.7 & & 16.3 & 16.9 & 11.0  & \textbf{51.6} \\
        Value misuse (26.8 \%)&40.9 & 50.0 & & 33.2 & 34.7 & 45.1  & \textbf{54.5} \\
        Operator misuse (15.2 \%)& 64.0 & 56.0 & & 32.8 & 42.0 & 34.4  & \textbf{68.0}  \\
        Variable misuse (14.0 \%)&30.4 & \textbf{43.5} & & 35.7  & 33.7 & 30.4  & \textbf{43.5}  \\
        Function misuse (4.9 \%)&37.5 & \textbf{50.0} & & 25.0 & 37.5 & 37.5  & 37.5  \\
        \midrule
        \textbf{Total}& 39.6 & 47.0 & & 28.1 & 30.4 & 31.8  & \textbf{51.2} \\
        \bottomrule
    \end{tabular}
}    
    \caption{
        Full results of error type analysis on HumanEvalFix with additional baselines. 
    }
    \label{table:error_type_full}
\end{table*}
\subsection{Benchmarks}\label{appendix:benchmarks}
For our experiments, we consider the following benchmarks:
\paragraph{HumanEvalPack. } 
HumanEvalPack is an extended version of HumanEval~\citep{chen2021codex} to evaluate code LLMs' instruction-following abilities onsix programming languages. HumanEvalPack consists of three tasks:
\begin{itemize}
    \item \textbf{HumanEvalSynthesize} corresponds to the original HumanEval benchmark used to assess code synthesis models. In this task, the models are prompted to synthesize the correct codes given a natural language (i) docstring or (ii) comment describing the desired code. 

    \item \textbf{HumanEvalFix} is a dataset that is manually curated using solutions from HumanEval for the task of code editing. 
    Given an (i) incorrect code function, which contains a subtle bug, and (ii) several unit tests (\ie, test cases), the model is tasked to correct/fix the function. 
    The dataset consists of 164 samples from the HumanEval solutions, and each sample comes with human-authored bugs across six different programming languages, thus covering 984 bugs in total. 
    The bugs are designed in a way that the code is executed without critical failure but fails to produce the correct output for at least one test case.
    
\end{itemize}
\paragraph{MBPP.}
MBPP dataset~\citep{austin2021program} is a benchmark designed to address entry-level programming problems in Python. 
The problems in this dataset range across diverse domains, such as programming fundamentals and standard library functionality. 
Each problem comprises a task description, code solution, and three test cases. 
The average, median, and maximum number of lines of the codes are 6.8, 5, and 50, respectively. Most of the problem descriptions are no longer that two sentences.
The dataset contains 974 samples in total, where 374, 90, 500 of them are for train, validation, test set, respectively.

\paragraph{APPS. }
APPS~\citep{hendrycksapps2021} consists of 10,000 coding problems, along with 232,421 human-authored golden solutions. The average length of a problem is around 293 words.
The dataset is split evenly into training and test sets. In the test set, each problem has 21.2 test cases for evaluating the solutions. This dataset has three categories of difficulty level:
\begin{itemize}
    \item The \textbf{introductory level} problems are designed to be solvable with simple algorithms for entry-level programmers with 1-2 years of experience in programming. For instance, counting the number of vowels in a string or returning the running sum of a list of integers. For this level, there are 2,639 and 1,000 problems in the training set and the test set, respectively.

    \item The \textbf{interview level} problems are designed to be more difficult such that they can be asked in programming technical interviews. These include questions on data structures such as trees or graphs, or questions on nontrivial algorithms. The numbers of problems at this level in the training/test sets are 2,000 and 3,000, respectively.
    
    \item The \textbf{competition level} problems are  the most challenging and require the level of advanced high school or collegiate programming competitions, including USACO, IOI, and ACM. 
    There are a total of 1,361 competition-level problems, 1,000 of which are in the test set.
\end{itemize}

\paragraph{Metrics. } We use pass@1 to measure the code editing performance for all benchmarks. Specifically, pass@1 is computed as the proportion of problems in the test set where the model correctly edits the incorrect solution:
\begin{equation}
    \text{Pass@1} = \frac{\text{(\#. problems with correct edits)}}{\text{(\#. problems in the test set)}} \times 100
\end{equation}

We additionally report ERR for code generation benchmarks to correctly assess the performance gain from code editing. In practice, we calculate ERR by dividing the gain in pass@1 from editing by 100 minus the pass@1 score before editing.
\begin{equation}
    \text{ERR} = \frac{\text{($\Delta$ Pass@1 after editing)}}{100-\text{(Pass@1 before editing)}} \times 100
\end{equation}
Note that ERR is introduced because there is a discrepancy between the evaluation methods for  code editing and generation benchmarks. Unlike code editing benchmarks, we only edit solutions that are found to be erroneous by running example test cases in code generation benchmarks. In MBPP, for example, 51\% of all test problems needed further code editing due to incorrect solutions generated from ChatGPT. Meanwhile, since all problems are given incorrect solutions in the code editing benchmark, ERR is equivalent to pass@1 for code editing benchmarks as pass@1 before editing is always 0. 

\paragraph{Terms and License. } Both APPS and HumanEvalPack datasets are licensed under the MIT License. MBPP dataset is licensed under Creative Commons by CC-BY 4.0. We have confirmed that all of the artifacts used in this paper are available for non-commercial, scientific use.

\subsection{Analysis on Train-test Overlap}\label{appendix:overlap}
\begin{figure*}[hbt!]
    \centering
        \begin{subfigure}[b]{0.3\textwidth}
        \includegraphics[width=\textwidth]{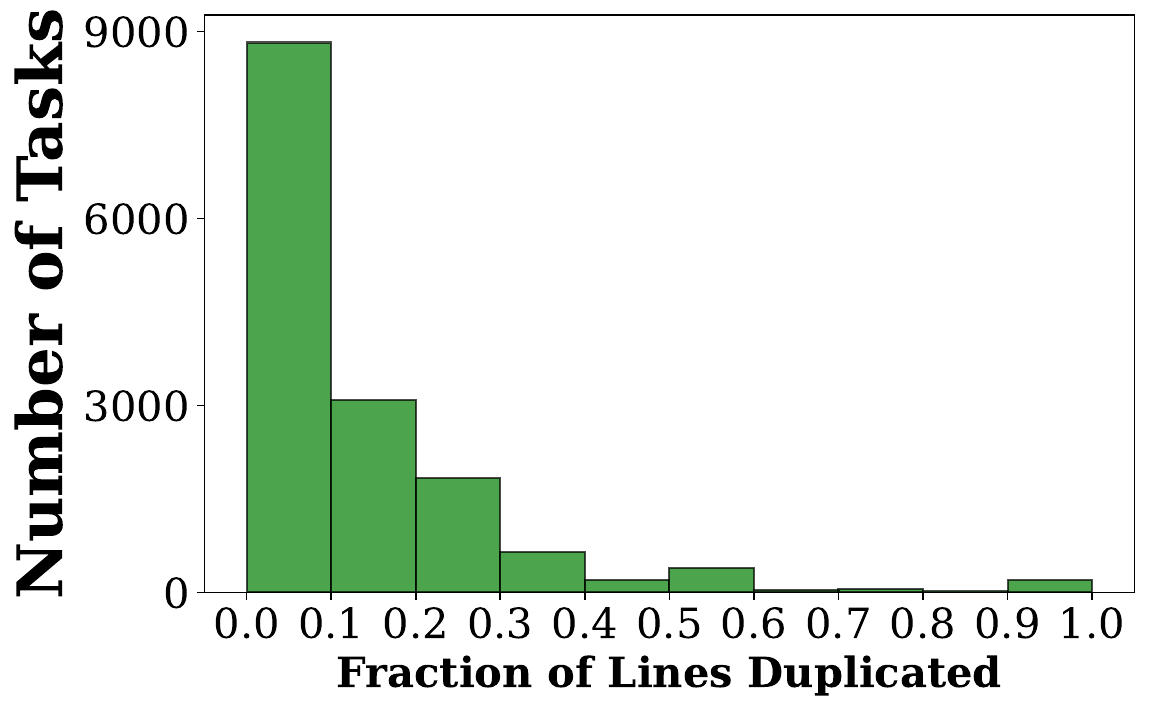}    
        \caption{APPS}
        \end{subfigure}
        \begin{subfigure}[b]{0.3\textwidth}
        \includegraphics[width=\textwidth]{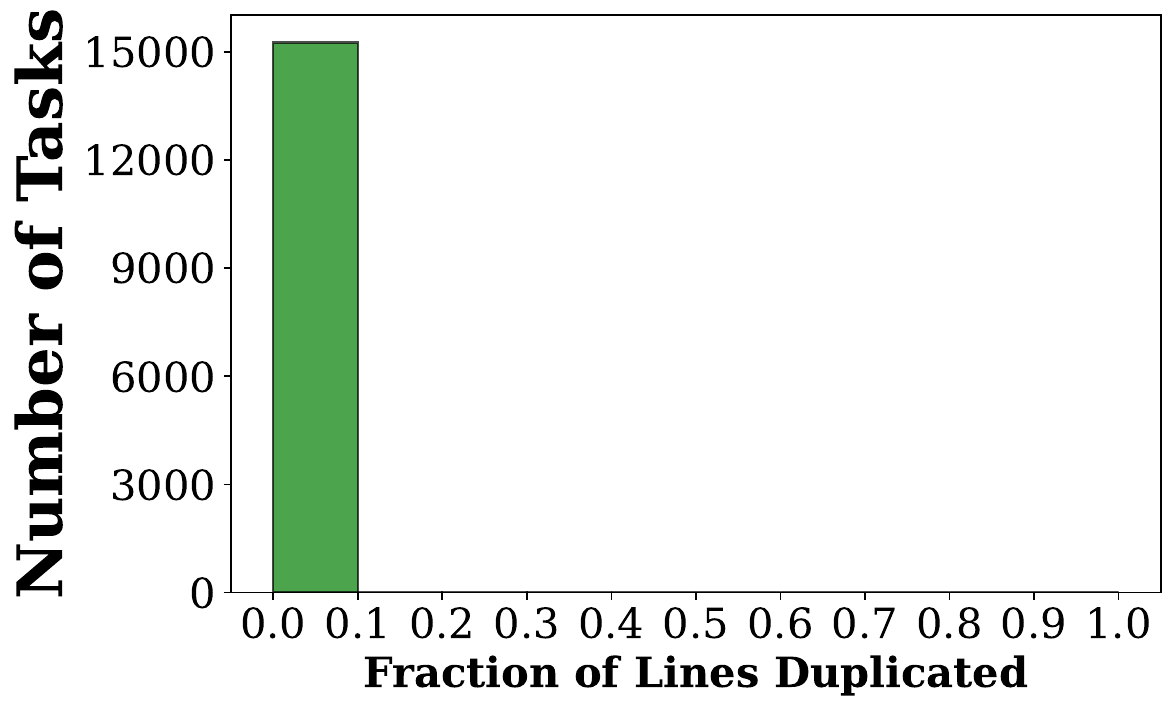}    
        \caption{HumanEvalPack}
        \end{subfigure}
        \begin{subfigure}[b]{0.3\textwidth}
        \includegraphics[width=\textwidth]{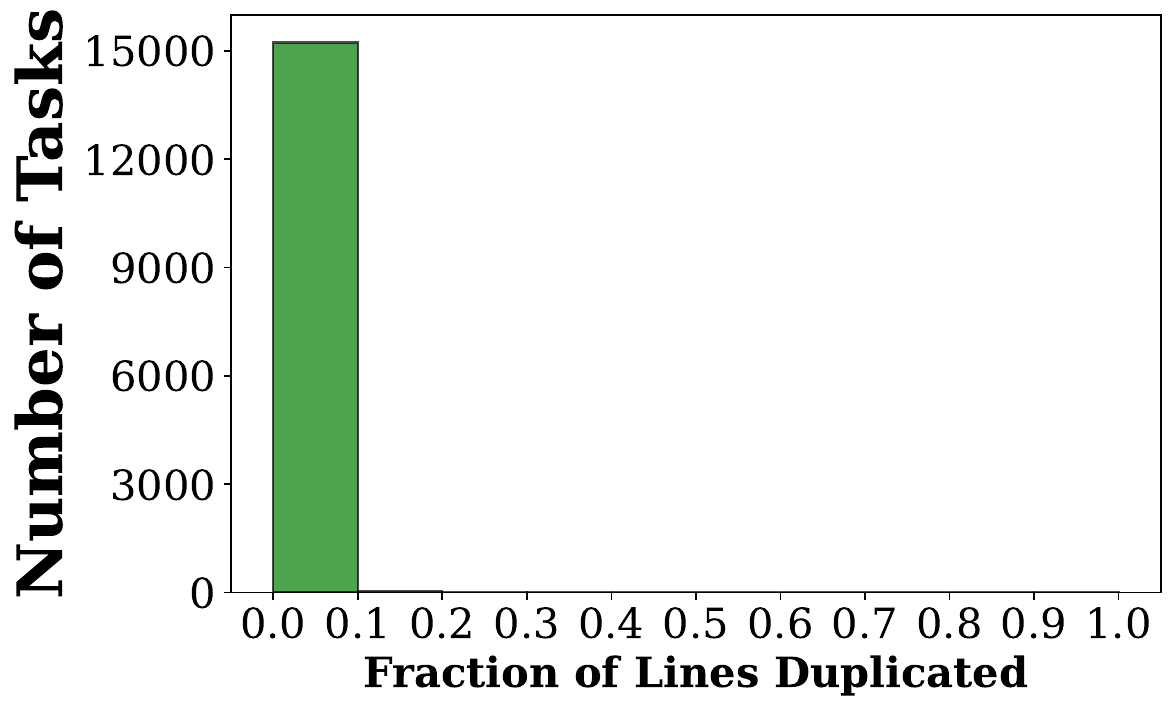}
        \caption{MBPP}    
        \end{subfigure}
    \caption{The fraction of line overlaps between \cf~and benchmark datasets, \ie, (a) APPS, (b) HumanEvalPack, and (c) MBPP.}
\label{fig:overlap-relative}
\end{figure*}

\begin{figure*}[hbt!]
    \centering
        \begin{subfigure}[b]{0.3\textwidth}
        \includegraphics[width=\textwidth]{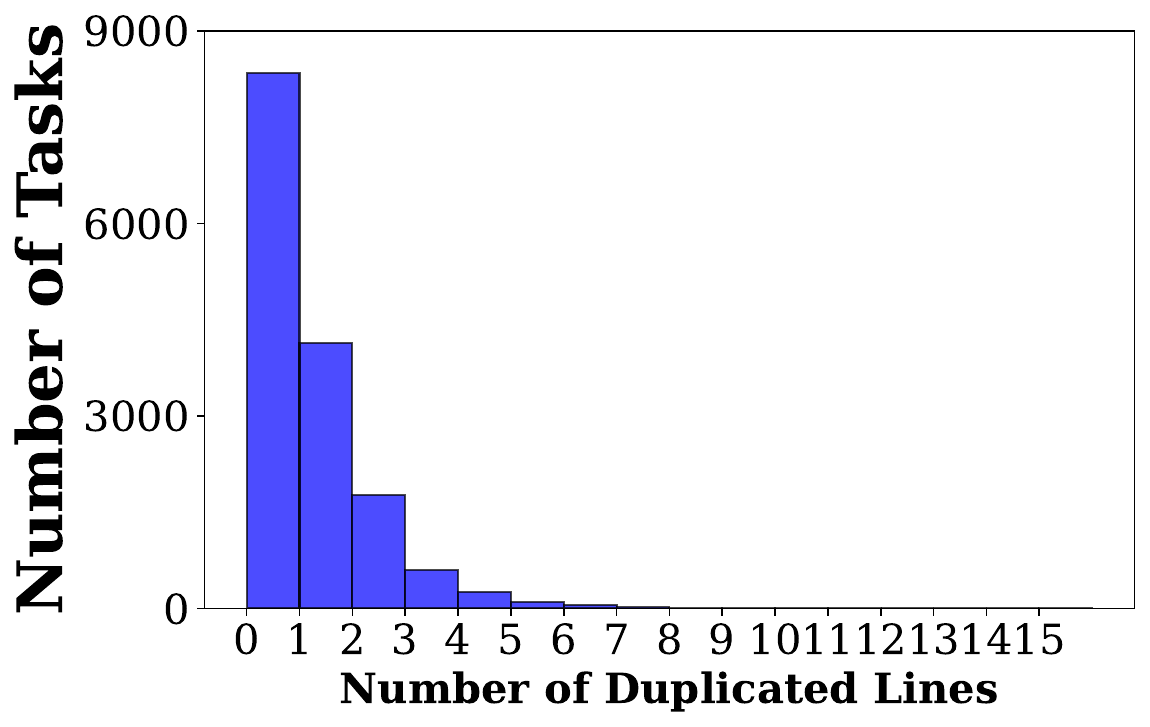}
        \caption{APPS}
        \end{subfigure}
        \begin{subfigure}[b]{0.3\textwidth}
        \includegraphics[width=\textwidth]{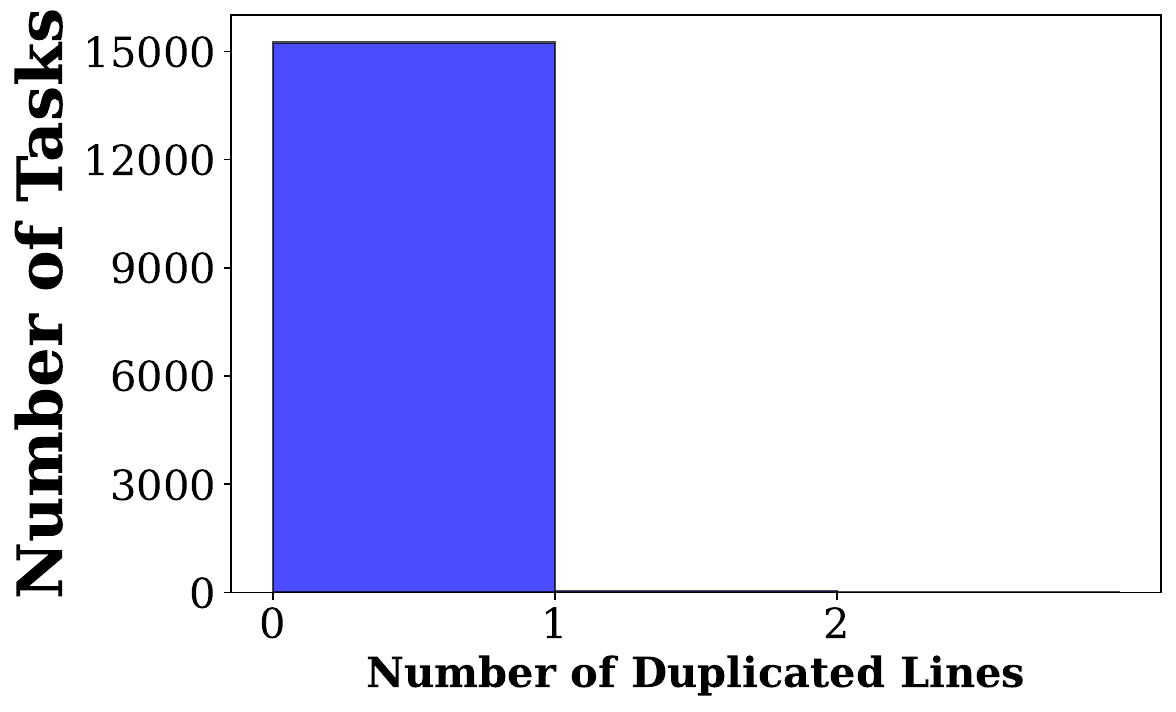}    
        \caption{HumanEvalPack}
        \end{subfigure}
        \begin{subfigure}[b]{0.3\textwidth}
        \includegraphics[width=\textwidth]{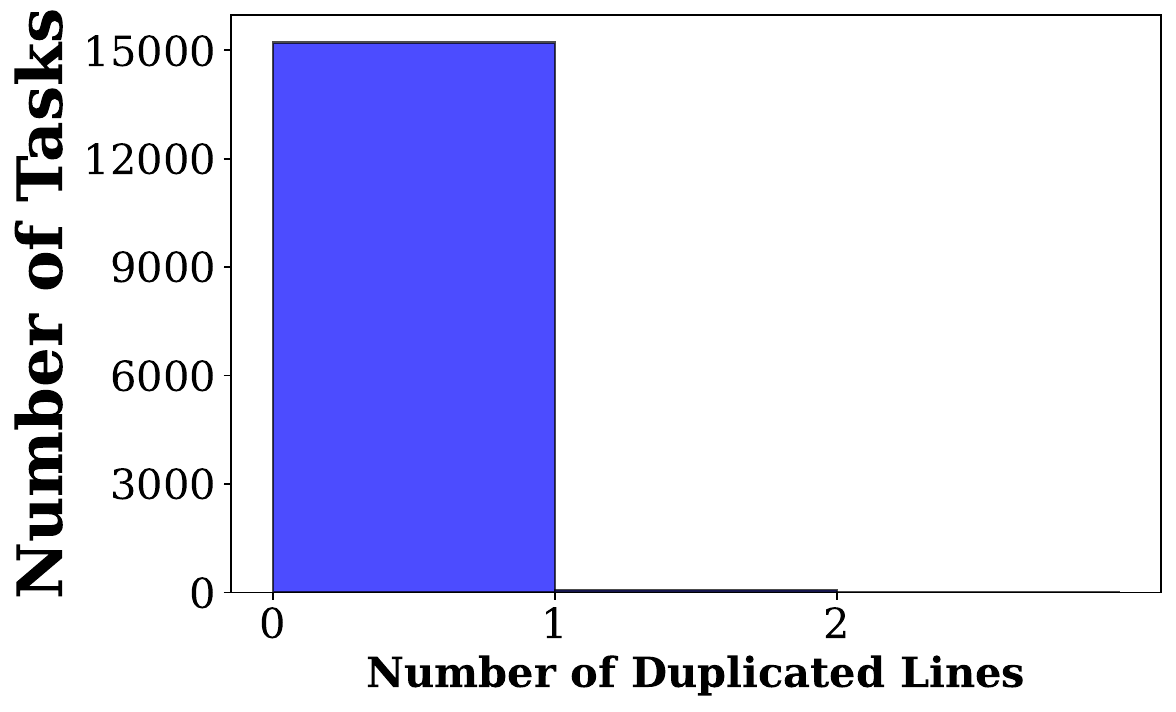}
        \caption{MBPP}   
        \end{subfigure}
    \caption{The absolute number of line overlaps between \cf~and benchmark datasets, \ie, (a) APPS, (b) HumanEvalPack, and (c) MBPP.}
\label{fig:overlap-absolute}
\end{figure*}

A possible concern is that the training data in \textsc{Coffee} might overlap with the test data in the adopted benchmarks (\ie, HumanEval, MBPS, and APPS). Therefore, we follow \citet{augustus2021program} and measure the amount of identical codes (based on the number of repeated lines) between the training and test data. Figure~\ref{fig:overlap-relative} reports the fraction of line overlaps, and Figure~\ref{fig:overlap-absolute} reports the absolute number of line overlaps.
We observe that most of solutions in \cf~do not contain lines that appear in the benchmark datasets which we evaluate our models on.
The results strongly suggests that both the critic and the editor do not solve the tasks by memorizing the dataset.


\subsection{Analysis on Different Error Types}\label{appendix:error_types}
We expand upon the analysis on error types in Table~\ref{table:error_type} and compare the performance of \textsc{CoffeePots} with additional open- and closed-source baselines, including ChatGPT, OctoGeeX, and WizardCoder. 
Table~\ref{table:error_type_full} shows that \textsc{CoffeePots} outperforms open-source baselines and ChatGPT for all error types, demonstrating its effectiveness in correcting different types of errors.


\subsection{Feedback Quality Evaluation}
\label{appendix:feedback_quality_eval}
For pairwise comparisons in Figure~\ref{fig:g-eval}, we use G-Eval~\citep{liu2023geval} and prompt \texttt{gpt-4-0613} to evaluate the feedback quality for the two inputs based on the following three criteria.
\begin{itemize}
    \item Identification: Does the feedback correctly identify errors in the code?
    \item Correction: Does the feedback provide a technically accurate explanation of the error?
    \item Validity: Is it possible to correct from incorrect code to correct code by using feedback?
\end{itemize}
GPT-4 is then prompted to choose the better feedback among the two candidates. We include the prompt used for the evaluation in Table~\ref{tab:gpt4_compare_prompt}.

For our analysis on feedback quality in Figure~\ref{fig:feedback_quality}, we prompt \texttt{gpt-4-0613} to rate the helpfulness of the feedback using 5-point Likert scale. Table~\ref{tab:gpt4_eval_prompt} shows the prompt used for our analysis, including detailed instructions on rating the helpfulness of feedback.

\subsection{Human Evaluation on Quality of Feedback}\label{appendix:human_error_detection}
\paragraph{Preparing feedback for the evaluation.}
We aim to analyze the quality of feedback on both the code editing benchmark (\ie{}, HumanEvalFix) and the code generation benchmarks (\ie{}, HumanEvalSynthesize, MBPP, and APPS).
For the code editing benchmark, we use the erroneous codes provided in the dataset.
However, as we do not have erroneous codes for the three code generation benchmarks, we use the codes generated by ChatGPT. Specifically, we curate the codes tested as incorrect by running test cases.
We randomly sample 100 codes from each the dataset to assure the correctness of our evaluation. 

\begin{figure*}[t]
\centering
    \includegraphics[width=0.92\linewidth]{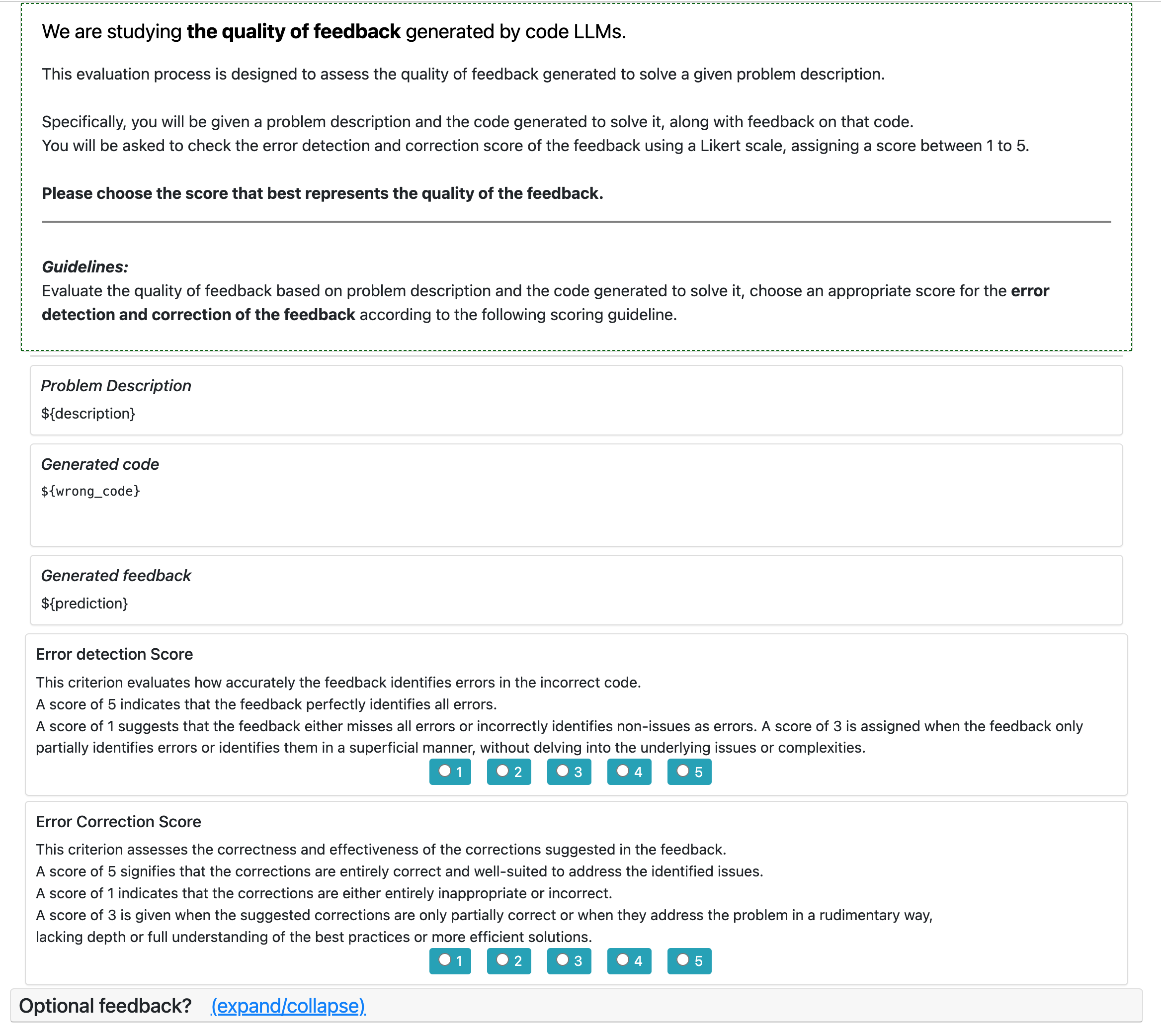}
\caption{The interface used for human evaluation on the quality of feedback generated from critic.} 
\label{fig:amt_prelim}
\end{figure*}

\begin{figure*}[t]
\centering
    \includegraphics[width=0.92\linewidth]{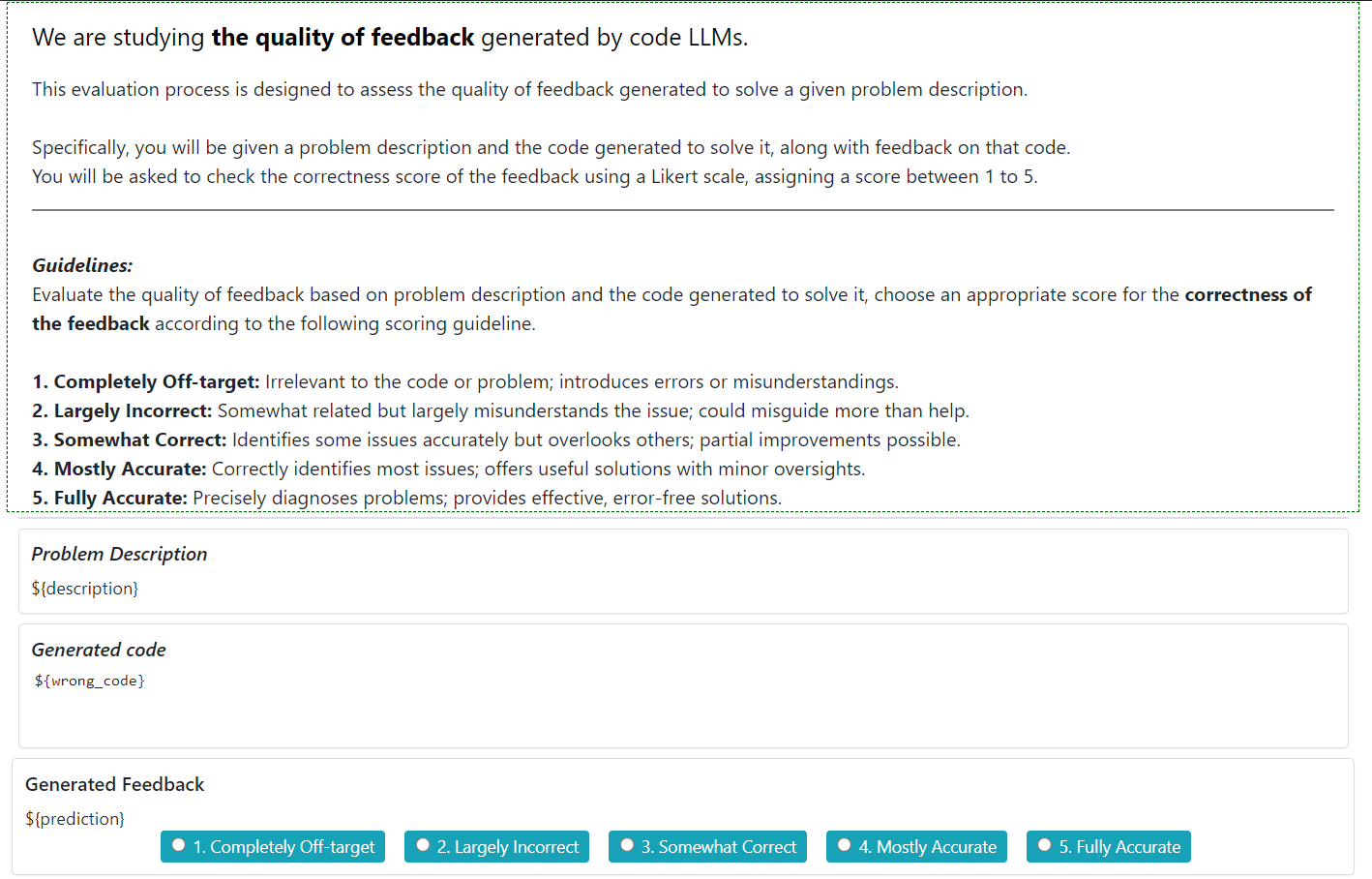}
\caption{The interface used for human evaluation on the correctness of feedback.} 
\label{fig:amt}
\end{figure*}

\begin{table}[t]
    \centering
    
    \resizebox{\linewidth}{!}{
    \begin{tabular}{lcccc}
    \toprule
         & \textbf{H.E.Fix} & \textbf{H.E.Synth} & \textbf{MBPP} & \textbf{APPS} \\
         \midrule
       Fleiss Kappa  & 0.332& 0.326& 0.342& 0.335\\
    \bottomrule
    \end{tabular}
    }
    \caption{Fleiss Kappa scores for human evaluation across each benchmark.}
    \label{tab:fleiss_kappa}
\end{table}
\paragraph{Details on human evaluation.}
We conduct human evaluation by using Amazon Mechanical Turk (AMT), which is a popular crowd sourcing platform.
As we need workers who have enough experience with Python, we conduct a qualification test to collect a pool of qualified workers.
In result, we recruit 63 workers who have passed the test, and task them to evaluate the quality of the feedback on Likert scale, ranging from 1 to 5.
Each sample is evaluated by three different raters to ensure the reliability.
Based on our estimates of time required per task, we ensure that
the effective pay rate is at least \$15 per hour.
For Figure~\ref{fig:prelim_error_analysis}, we use the evaluation interface in Figure~\ref{fig:amt_prelim}. For Table~\ref{tab:error_detection_and_err}, we use the evaluation interface in Figure~\ref{fig:amt}.
We also report the Fleiss kappa~\citep{fleiss1971measuring} in Table~\ref{tab:fleiss_kappa} to evaluate the consistency of the evaluation among annotators.

\begin{table}[t]
\small
\renewcommand{\arraystretch}{1.15} 
\centering

\begin{tabular}{lc}
\toprule
\textbf{Resource of feedback}&\textbf{Pass@1}\\
\midrule
Preference-tuned critic (\cref{ssec:dpo})  &  45.9\\
Random in-batch feedback & 25.6  \\
GPT-4  & \textbf{61.7}    \\

\bottomrule
\end{tabular}

\caption{The performances of our editors when augmented with different types of feedback (Pass@1) on the HumanEvalFix benchmark.}
\label{tab:critic_analysis}
\end{table}

\begin{table}[t]
\small
\renewcommand{\arraystretch}{1.15} 
\centering

\begin{tabular}{lc}
\toprule
\textbf{Approach of Feedback Selection}&\textbf{Pass@1}\\
\midrule
Feedback selector (\cref{ssec:feedback_selector}) & 45.1  \\
Random selection  & 39.6 \\
Oracle selector  & \textbf{66.4} \\

\bottomrule
\end{tabular}

\caption{The performance of our editors when applying different approaches of feedback selection (pass@1) on HumanEvalFix.}
\label{tab:selector_analysis}
\end{table}

\begin{table}[t]
    \centering
    \small
    \resizebox{\columnwidth}{!}{
    \begin{tabular}{lccccc}

    \toprule
    \textbf{\textit{K}} & 1 & 3 & 5 & 7 & 9  \\ 
    \midrule
    \textbf{Pass@1} & 45.9 & 48.7 & \textbf{51.2} & 49.4 & 48.1 \\
    \bottomrule
    \end{tabular}
    }
    \caption{
        Anaylsis of diverse number of sampling $K$.
    }
    \label{tab:k_analysis}
\end{table}
\subsection{Analyses on Different Feedback Types and Selection Strategies}
To further analyze the effect of feedback and feedback selection, we conduct experiments of code editing on the HumanEvalFix benchmark using various types of feedback and feedback selection.

\paragraph{Feedback types.} We compare the performance of our editor when given different types of feedback: (1) feedback generated by preference-tuned critic; (2) feedback generated by GPT-4; (3) random feedback for other problem $q$ within the same batch, which is also generated by preference-tuned critic.
Table~\ref{tab:critic_analysis} shows that there is a performance gain when the editors are given feedback from preference-tuned critic compared to random in-batch feedback. This demonstrates the effectiveness of our approach. However, there remains a gap between the improvements from feedback from our preference-tuned critic and feedback from GPT-4.

\paragraph{Selection strategies.} Here, we compare three settings of feedback selection: (1) Our feedback selector $\psi$;
(2) An oracle selector. We simulate the oracle selector by first over-sampling feedback from critic and selecting the one that yields the desired output for the given problem; (3) Random selection from over-sampled feedback. 
Table~\ref{tab:selector_analysis} shows that oracle feedback leads to a large performance gain compared to randomly selected feedback. This supports our hypothesis that helpful feedback improves the editor's performance in code editing.

\subsection{Ablation on the Number of Sampled Feedback ($K$)}
\label{app:k_analysis}
We also ablate the number of sampled feedback in feedback selection, denoted as $K$, to examine its impact on the performance. As shown in Table~\ref{tab:k_analysis}, the setting under $K=5$ shows the best results among 5 different parameters, \ie{}, $K = 1, 3, 5, 7, 9$, validating the parameter choice of our setting.

\begin{figure}[t]
\centering
    \includegraphics[width=0.9\linewidth]{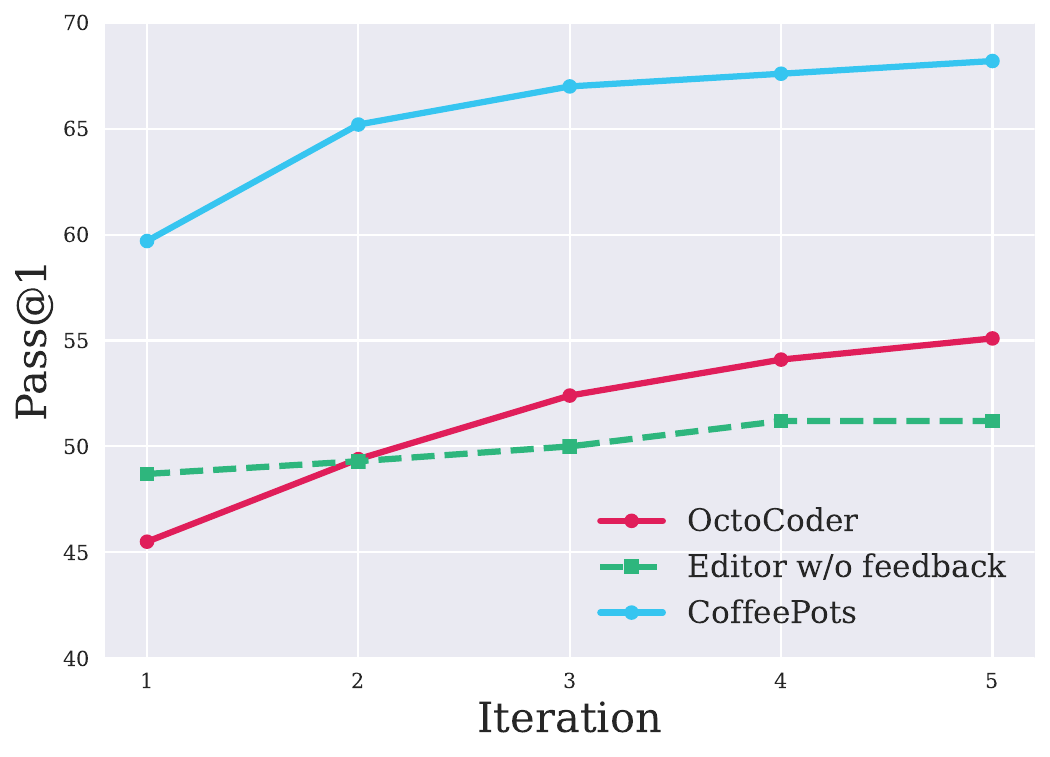}
\caption{Performance on test cases from HumanEvalFix, measured under the iterative edit setting.}
\label{fig:iteration}
\end{figure}

\subsection{Iterative editing.}
Inspired by \citet{zhang2023selfedit}, we consider a practical setting where the models are tasked to edit codes iteratively. 
Specifically, we task the models to refine the code iteratively by testing the edited codes with sample test cases and selecting wrong solutions for further editing in the next iteration.\footnote{Each problem in HumanEvalFix has two separated test case sets and we use the set is accessible at inference.}
Figure~\ref{fig:iteration} compares the performances of our editor without feedback, our editor with feedback (\ie, \textsc{CoffeePots}), and baselines over five iterations. 
We first observe that the editing performance of \textsc{CoffeePots} not only consistently increases over iterations but also outperforms OctoCoder in every iteration.
However, when feedback is absent, the performance gains from the editor are marginal compared to \textsc{CoffeePots}.
We hypothesize that such suboptimal improvement may be due to the additional errors introduced by the editor, as it lacks guidance from feedback. 

\section{Case Study}\label{appendix:case_study}

\begin{figure*}[ht]
\centering
    \includegraphics[width=1.3\columnwidth]{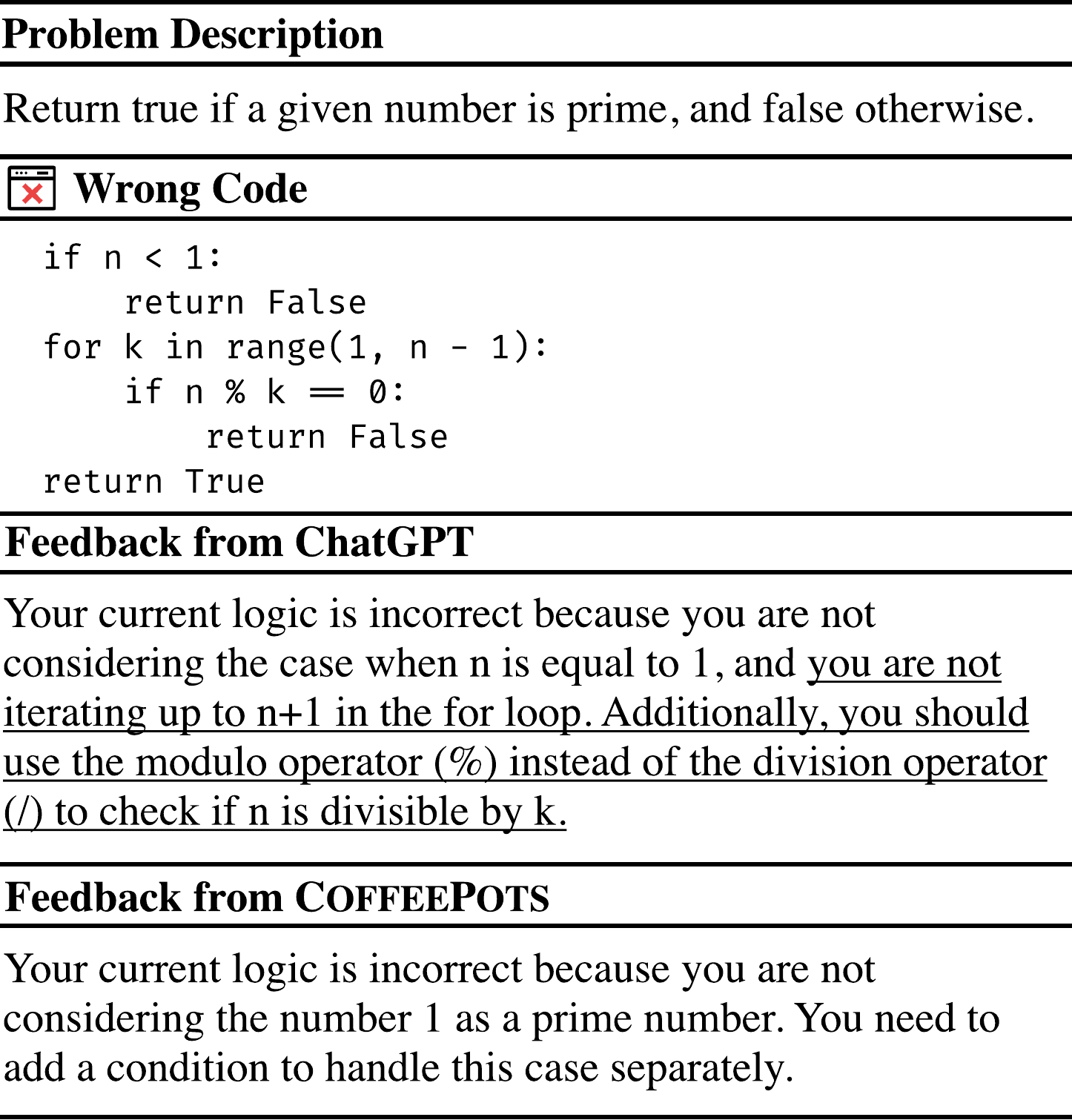}
\caption{Examples of the feedback from ChatGPT and the feedback from \cpemoji~\textsc{CoffeePots}.}
\label{fig:case_study_feedback}
\end{figure*}

In Figure~\ref{fig:case_study_feedback}, we present examples of generated feedback.
Although ChatGPT correctly identifies the incorrect parts in the wrong code, it also provides unnecessary feedback (underlined), which may confuse the editor in feedback-augmented code editing.
In contrast, \cpemoji~\textsc{CoffeePots} provides helpful feedback on the incorrect part without unnecessary information.

\begin{figure*}[ht]
\centering
    \includegraphics[width=1.95\columnwidth]{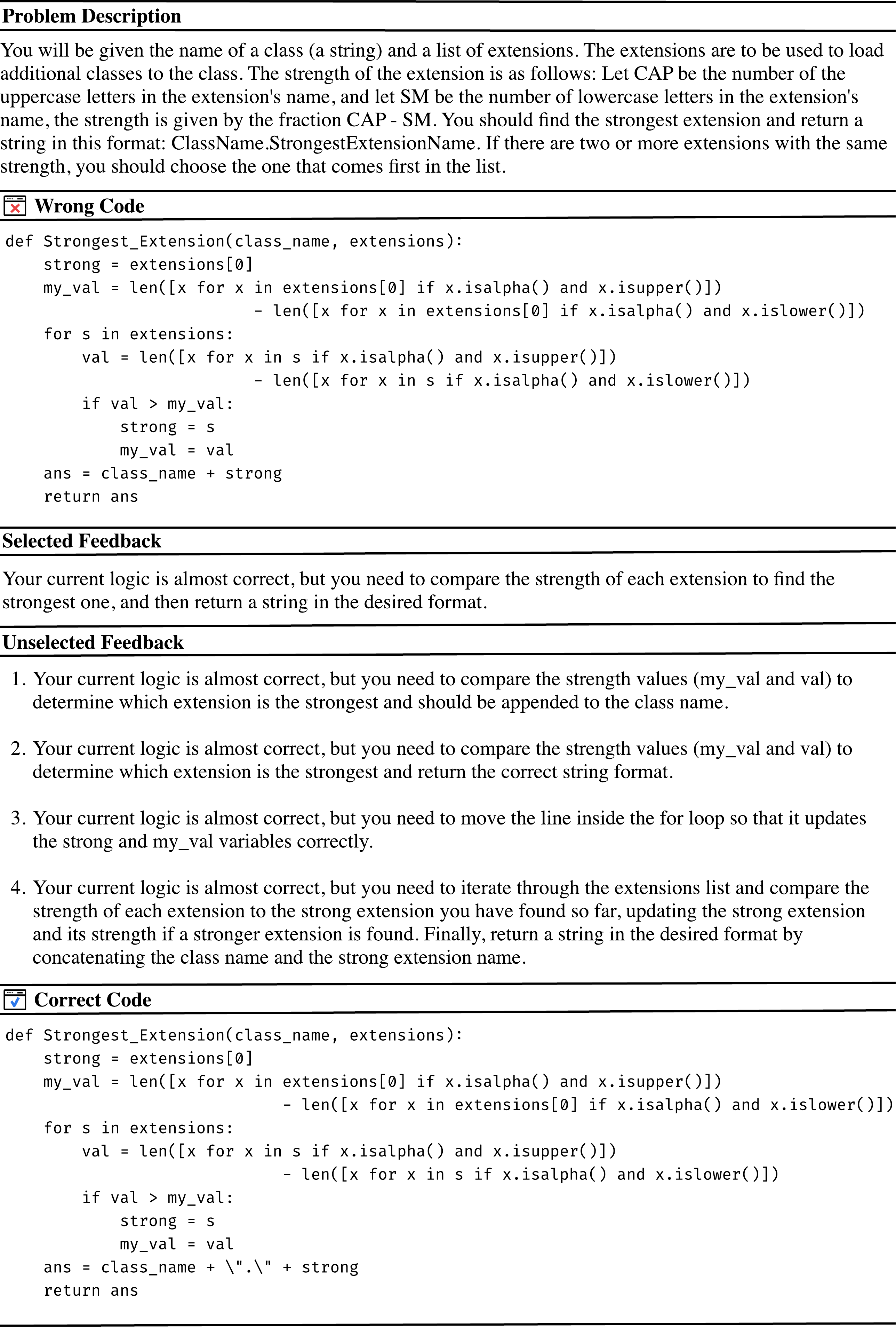}
\caption{Example of the generation and selection of feedback and the editing from \cpemoji~\textsc{CoffeePots}.}
\label{fig:case_study_coffeepots}
\end{figure*}

We also conduct an empirical case study to assess how well the feedback selector filters out inappropriate feedback.
In Figure~\ref{fig:case_study_coffeepots}, the wrong code exhibits an issue where the final output does not match the desired format.
Among the unselected feedback, some fails to identify format error (\eg, case 1, 3), while others contain a lot of unnecessary information (\eg, case 4).
However, the feedback selector demonstrates its ability to choose useful feedback, leading to the correct code.


\begin{table*}[t]
    \centering
    \small
    \begin{tabular}{p{14cm}}
        \toprule
        \textbf{Input} \\
        \midrule
        Provide feedback on the errors in the given code\\
        \\
        \textbf{Problem Description:} \\
        \{description\} \\
        \\
        \textbf{Incorrect Code:} \\
        \{incorrect\_code\} \\
        \\
        \toprule
        \textbf{Output} \\
        \midrule
        \textbf{Feedback:} \\
        \{feedback\} \\
        \bottomrule
    \end{tabular}
    
    \caption{The input-output format of the critic model.}
    \label{tab:critic_prompt}
\end{table*}
\begin{table*}[t]
    \centering
    \small
    \begin{tabular}{p{14cm}}
        \toprule
        \textbf{Input} \\
        \midrule
        Provide feedback on the errors in the given code and suggest the correct code to address the described problem.\\
        \\
        \textbf{Problem Description:} \\
        \{description\} \\
        \\
        \textbf{Incorrect Code:} \\
        \{incorrect\_code\} \\
        \\
        \textbf{Feedback:} \\
        \{feedback\} \\
        \\
        \toprule
        \textbf{Output} \\
        \midrule
        \textbf{Correct code:} \\
        \{correct\_code\} \\
        \bottomrule
    \end{tabular}
    
    \caption{The input-output format of the editor model.}
    \label{tab:editor_prompt}
\end{table*}
\begin{table*}[t]
    \centering
    \small
    \begin{tabular}{p{14cm}}
        \toprule
        \textbf{Prompt} \\
        \midrule
        Generate feedback that guides the refinement from Incorrect Code to Correct Code. The feedback is not about the superficial style of the code, but about the internal logic of the code. So do not mention about the changes of variable names or the format of the code. Please point out the incorrect logic of the code and provide succinct, constructive feedback in one sentence. And please do not mention about the correct code in feedback. \\
        \\
        \texttt{- Example 1 -} \\
        \textbf{Problem Description:} Tom is currently at point (x, y). The rectangle has sides parallel to the coordinate axes, with the bottom left vertex at (0, 0) and the top right vertex at (w, h). Write a program to find the minimum distance to the boundary of the rectangle. \\
        \textbf{Incorrect Code}: \\
        x, y, w, h = map(int, input().split())\\
        if w-x < x or h-y < y:\\
            \hspace{1cm}if w-x < h-y:\\
                \hspace{2cm}print(w-x)\\
            \hspace{1cm}else:\\        
                \hspace{2cm}print(h-y)\\
        elif w-x > x or h-y > y:\\
            \hspace{1cm}if x < y:\\
                \hspace{2cm}print(x)\\
            \hspace{1cm}else:\\
                \hspace{2cm}print(y)\\
        \textbf{Correct Code:} \\
        x, y, w, h = map(int, input().split())\\
        resultY = min(h -y, y)\\
        resultX = min(w - x, x)\\
        if resultX < resultY:\\
            \hspace{1cm}print(resultX)\\
        else:\\
            \hspace{1cm}print(resultY)\\
        \textbf{Feedback for Refining the Code:} The logic in your if-statements is flawed, as you need to first independently identify the smallest distances from point (x, y) to the lines x=0, y=0, w, and h, and then find the minimum amongst them.\\
        \\ 
        \texttt{- Example 2 -} \\ 
        \textbf{Problem Description:} Three integers A, B, and C are given. Write a program that prints the second largest integer.\\
        \textbf{Incorrect Code:} \\
        A, B, C = list(map(int,input().split()))\\
        print(max(A,B,C)) - print(min(A,B,C))\\
        Correct Code: \\
        data = []\\
        a,b,c = map(int,input().split())\\
        data.append(a)\\
        data.append(b)\\
        data.append(c)\\
        data.sort()\\
        print(data[1])\\
        \textbf{Feedback for Refining the Code:} Your current logic is incorrect because subtracting the smallest integer from the largest doesn't necessarily give you the second largest integer. Instead, you should add the three integers to a list, sort the list, and print the second element.\\
        \\
        \texttt{- Example 3 -} \\ 
        \textbf{Problem Description: }..... \\
        \textbf{Incorrect Code:} ... \\
        \\
        
        \texttt{- Example 4 -} \\ 
        \textbf{Problem Description:} \\
        \{problem description\}
        \\
        \textbf{Incorrect Code:} \\
        \{incorrect code\}
        \\
        \textbf{Correct Code:} \\
        \{correct code\}
        \\
        \textbf{Feedback for Refining the Code:} \\
        \bottomrule
    \end{tabular}
    
    \caption{The prompt for generating feedback. We prompt ChatGPT to generate feedback in a 3-shot setting (Example 3 is omitted in this table, due to limited space).}
    \label{tab:prompt-annotation}
\end{table*}
\begin{table*}[t]
    \centering
    \small
    \begin{tabular}{p{14cm}}
        \toprule
        \textbf{Prompt} \\
        \midrule
        Given the input format and python code, please provide at least 30 challenging test input values to evaluate its functionality.\\
        For every start of samples, please attach <start> token to indicate that the input string has started. \\
        Also, for every end of samples, please attach <end> token to indicate that the input string has ended.\\
        \\
        \textbf{input format:} \\
        \{input format\}
        \\
        \textbf{python code:} \\
        \{python code\}
        \\
        \textbf{Sample:} \\
        \bottomrule
    \end{tabular}
    
    \caption{The prompt for generating synthetic test cases. We prompt ChatGPT to generate hidden test cases for given Python code.}
    \label{tab:synthetic_test_case}
\end{table*}

\begin{table*}[t]
    \centering
    \small
    \begin{tabular}{p{14cm}}
        \toprule
        \textbf{Prompt} \\
        \midrule
        Consider the following aspects when evaluate the feedback:\\
        \\
        Identification: Does the feedback correctly identify the error in the code?\\
        \\
        Correction: Does the feedback provide a technically accurate explanation of the error?\\
        \\
        Validity: Is it possible to correct from incorrect code to correct code by using feedback?\\
        \\
        Score each aspect from 1 to 5, with 1 being the lowest and 5 the highest. Then, calculate the average score to determine the overall effectiveness of the feedback.\\
        \\
        \textbf{Problem Description:}\\
        \{problem\}\\
        \\
        \textbf{Incorrect Code:} \\
        \{incorrect\_code\} \\
        \\
        \textbf{First feedback:}\\
        \{feedback\_1\}\\
        \\
        \textbf{Second feedback:}\\
        \{feedback\_2\}\\
        \\
        \textbf{Correct code:}\\
        \{correct\_code\}\\
        \\
        Generate your assessment in the following format:\\
        \\
        Assessment: [Assessment]\\
        \\
        INDEX: [Index] (Use A for select first feedback, B for select second feedback, or N if neither is superior)\\
        \bottomrule
    \end{tabular}
    
    \caption{The prompt for GPT-4 evaluation. We prompt GPT-4 to compare the feedback quality on coding errors, focusing on clarity, accuracy, and validity.}
    \label{tab:gpt4_compare_prompt}
\end{table*}
\begin{table*}[t]
    \centering
    \small
    \begin{tabular}{p{14cm}}
        \toprule
        \textbf{Prompt} \\
        \midrule
        You will be provided with a description of a problem, wrong code, and the correct code that is intended to solve the issue.\\
        \\
        Along with these, you will receive a feedback addressing the inaccuracies or issues in the wrong code.\\
        \\
        Your task is to carefully evaluate feedback.\\
        \\
        Evaluation Criteria:\\
        \\
        Helpfulness (1-5): Evaluate the extent to which the feedback is helpful by providing a clear and constructive direction that leads to the accurate correction of the given code.\\
        \\
        The assessment should focus on the relevance, specificity, and practicality of the feedback and its effectiveness in guiding the correct revision. The scale consists of: \\
        \\
        1. Unhelpful: The feedback either misinterprets the code's issue or provides information unrelated to the problem, offering no constructive advice or path toward correction. \\
        \\
        2. Marginally Helpful: The feedback identifies an aspect of the code's problem but is vague or imprecise, providing limited guidance that could nudge the revision in the right direction, with little assurance of complete correction. \\
        \\
        3. Helpful: The feedback recognizes an issue and suggests a general course of action that could lead to improvements in the code. Yet, it may lack the level of detail needed to guarantee a thorough and accurate fix. \\
        \\
        4. Quite Helpful: The feedback clearly identifies key problems and provides a specific recommendation that aligns well with the necessary corrections, likely resulting in a proper fix for the primary issue if followed. \\
        \\
        5. Very Helpful: The feedback succinctly pinpoints a critical issue and offers a clear, focused recommendation that, if implemented, would significantly enhance the code's performance. \\
        \\
        For each feedback instance, provide a detailed assessment of its helpfulness and overall direction toward a solution, and assign an appropriate score under the FINAL\_SCORE token at the end.\\
        \\
        \textbf{Problem Description:}\\
        \{problem\}\\
        \\
        \textbf{Incorrect Code:} \\
        \{incorrect\_code\} \\
        \\
        \textbf{Correct Code:}\\
        \{correct\_code\}\\
        \\
        \textbf{Feedback:}\\
        \{Feedback\}\\
        \\
        \textbf{Feedback Evaluation:}\\
        \bottomrule
    \end{tabular}
    
    \caption{The prompt for G-Eval. We prompt GPT-4 to evaluate the feedback quality on coding errors, focusing on helpfulness.}
    \label{tab:gpt4_eval_prompt}
\end{table*}
\begin{table*}[t]
    \centering
    \small
    \begin{tabular}{p{14cm}}
        \toprule
        \textbf{Prompt} \\
        \midrule
        In this task, you are presented with a coding problem, including examples of both incorrect and correct code. You will also receive feedback critiquing the incorrect code. Your primary responsibility is to evaluate this feedback based on the following criteria.\\
        \\
        \textbf{Evaluation Criteria}:\\
        1. Error Identification (Score: 0.0 to 1.0): This criterion evaluates how accurately the feedback identifies errors in the incorrect code. A score of 1.0 indicates that the feedback perfectly identifies all errors. A score of 0.0 suggests that the feedback either misses all errors or incorrectly identifies non-issues as errors. A score of 0.5 is assigned when the feedback only partially identifies errors or identifies them in a superficial manner, without delving into the underlying issues or complexities.\\
        \\
        2. Correction Quality (Score: 0.0 to 1.0): This criterion assesses the correctness and effectiveness of the corrections suggested in the feedback. A score of 1.0 signifies that the corrections are entirely correct and well-suited to address the identified issues. A score of 0.0 indicates that the corrections are either entirely inappropriate or incorrect. A 0.5 score is given when the suggested corrections are only partially correct or when they address the problem in a rudimentary way, lacking depth or full understanding of the best practices or more efficient solutions. \\
        \\
        \textbf{Evaluation Steps:}\\
        1. Review All Materials: \\
        - Read the problem description, incorrect code, correct code, and the provided feedback.\\
        2. Evaluate Error Identification:\\
        - Determine if the feedback correctly points out errors in the incorrect code.\\
        - Assign a score from 0.0 (no correct identification) to 1.0 (all errors correctly identified).\\
        3. Evaluate Correction Quality:\\
        - Assess if the suggested corrections in the feedback are appropriate and correct.\\
        - Assign a score from 0.0 (completely inappropriate or incorrect corrections) to 1.0 (perfectly appropriate and correct).\\
        \\
        \textbf{Scoring:}\\
        Provide a float score between 0.0 and 1.0 for each criterion based on your evaluation.\\
        \\
        \textbf{Problem Description:}\\
        \{problem\}\\
        \\
        \textbf{Incorrect Code:} \\
        \{incorrect\_code\} \\
        \\
        \textbf{Correct Code:}\\
        \{correct\_code\}\\
        \\
        \textbf{Feedback:}\\
        \{Feedback\}\\
        \\
        \textbf{Evaluation Form (scores ONLY):}\\
        1. Error Identification: [Score]\\
        2. Correction Quality: [Score]\\
        \bottomrule
    \end{tabular}
    
    \caption{The prompt used in error analysis. We prompt ChatGPT to evaluate the feedback quality in error detection and error correction.}
    \label{tab:error_cor_det}
\end{table*}

\end{document}